\pdfoutput=1

\documentclass[11pt]{article}

\usepackage{EMNLP2023}

\usepackage{times}
\usepackage{makecell}
\usepackage[inline]{enumitem} 
\newlist{inparaenum}{enumerate*}{1}
\usepackage{listings}
\usepackage[frozencache,cachedir=.]{minted}

\usepackage{latexsym}
\usepackage{algorithm,algpseudocode}
\newcommand{\emnlpcr}[1]{{#1}}

\usepackage{amsmath}
\usepackage{amssymb}
\usepackage{float}
\usepackage{graphicx}
\usepackage{booktabs}
\usepackage{multirow}
\usepackage{hyperref}
\usepackage{ulem}
\usepackage{xspace}
\usepackage{pgfplots}
\usepackage[utf8]{inputenc}
\DeclareUnicodeCharacter{2212}{−}
\usepgfplotslibrary{groupplots,dateplot}
\usetikzlibrary{patterns,shapes.arrows}
\pgfplotsset{compat=newest}
\usepackage{lipsum,adjustbox}
\usepackage{subfigure}
\usepackage{url}

\usepackage{minitoc}

\definecolor{deepblue}{rgb}{0,0,0.5}
\definecolor{deepred}{rgb}{0.6,0,0}
\definecolor{deepgreen}{rgb}{0,0.5,0}
\definecolor{darkgreen}{RGB}{43,163,39}
\definecolor{bluesquare}{rgb}{126,166,224}
\definecolor{LightGray}{gray}{0.9}
\definecolor{DarkGray}{gray}{0.1}
\usepackage{minted}

\usepackage{tcolorbox}
\tcbuselibrary{minted,skins}

\newtcblisting{mycode}{
    listing engine=minted,
    colback=white,
    colframe=gray!75!black,
    listing only,
    minted style=default,
    minted language=python,
    minted options={
        fontsize=\small,
        numbersep=5pt,
        breaklines,
        breakanywhere,
    },
    left=1mm,
    enhanced,
    overlay={
        \begin{tcbclipinterior}
            \fill[gray!25] (frame.south west) rectangle ([xshift=1mm]frame.north west);
        \end{tcbclipinterior}
    },
    boxrule=0.5pt
}

\definecolor{codegreen}{rgb}{0,0.6,0}
\definecolor{codegray}{rgb}{0.5,0.5,0.5}
\definecolor{codepurple}{rgb}{0.58,0,0.82}
\definecolor{backcolour}{rgb}{0.95,0.95,0.92}

\usepackage[T1]{fontenc}

\usepackage[utf8]{inputenc}

\usepackage{microtype}

\usepackage{inconsolata}

\title{Let's Sample Step by Step:\\ \ours for Efficient Reasoning and Coding with LLMs}

\author{Pranjal Aggarwal\textsuperscript{1} \quad Aman Madaan\textsuperscript{3} \quad Yiming Yang\textsuperscript{3} \quad Mausam\textsuperscript{1,2} \\
        \textsuperscript{1}Department of Computer Science, Indian Institute of Technology, Delhi \\
        \textsuperscript{2}Yardi School of Artificial Intelligence, Indian Institute of Technology, Delhi \\
        \textsuperscript{3}Language Technologies Institute, Carnegie Mellon University \\
        {\normalsize \texttt{\href{mailto:pranjal2041@gmail.com}{pranjal2041@gmail.com}, \href{mailto:amadaan@cs.cmu.edu}{amadaan@cs.cmu.edu}, \href{mailto:yiming@cs.cmu.edu}{yiming@cs.cmu.edu},
         \href{mailto:mausam@cse.iitd.ac.in}{mausam@cse.iitd.ac.in}}}}

\newcommand{\aman}[1]{\textcolor{red}{[Aman: #1]}\xspace}

\newcommand{\todo}[1]{\textcolor{red}{[\textcolor{blue}{TODO:} #1]}}

\newcommand{\cmmnt}[1]{\ignorespaces}

\newcommand{\self}[0]{Self-Consistency\xspace}
\newcommand{\adaptive}[0]{Adaptive-Consistency\xspace}
\newcommand{\ours}{Adaptive-Consistency\xspace}

\newcommand{\stopfunc}[0]{stopping criteria}
\newcommand{\Stopfunc}[0]{Stopping Criteria}

\newcommand{\tablestd}[1]{{\tiny \textcolor{gray}{$(\pm #1)$}}}

\newcommand{\boost}[1]{$#1 \times$}

\newcommand{\codex}{\textsc{code-davinci-002}\xspace}
\newcommand{\incoder}{\textsc{Incoder-6B}\xspace}
\newcommand{\codegen}{\textsc{CodeGen-16B}\xspace}

\newcommand{\chatgpt}{\textsc{GPT-3.5-turbo}\xspace}
\newcommand{\vicuna}{\textsc{Vicuna-13B}\xspace}

\newcommand{\mathematical}{\textsc{Math}\xspace}
\newcommand{\commonsense}{\textsc{Commonsense}\xspace}
\newcommand{\symbolic}{\textsc{Symbolic}\xspace}
\newcommand{\coding}{\textsc{Code Generation}\xspace}

\newcommand{\pal}{PAL}

\newcommand{\Cot}{\textsc{Chain of Thought}\xspace}
\newcommand{\CoT}{\textsc{CoT}\xspace}
\newcommand{\llms}{LLMs\xspace}
\newcommand{\llm}{LLM\xspace}

\newcommand{\strategyqa}{\textsc{StrategyQA}\xspace}
\newcommand{\dateunderstanding}{\textsc{Date Understanding}\xspace}
\newcommand{\svamp}{\textsc{SVAMP}\xspace}
\newcommand{\asdiv}{\textsc{ASDIV}\xspace}
\newcommand{\gsm}{\textsc{GSM-8K}\xspace}
\newcommand{\snarks}{\textsc{Snarks}\xspace}
\newcommand{\salienttranslation}{\textsc{Salient Translation}\xspace}
\newcommand{\trackingthree}{\textsc{Tracking Shuffled Objects}\xspace}
\newcommand{\logicalthree}{\textsc{Logical Deduction}\xspace}
\newcommand{\ruinnames}{\textsc{Ruin Names}\xspace}
\newcommand{\disambiguationqa}{\textsc{Disambiguation QA}\xspace}
\newcommand{\booleanexpressions}{\textsc{Boolean Expressions}\xspace}
\newcommand{\penguins}{\textsc{Penguins}\xspace}

\newcommand{\humaneval}{\textsc{HumanEval}\xspace}
\newcommand{\mbpp}{\textsc{MBPP}\xspace}
\newcommand{\apps}{\textsc{APPS}\xspace}
\newcommand{\codecontests}{\textsc{CodeContests}\xspace}

\newcommand{\random}{\textsc{Random}\xspace}
\newcommand{\majority}{\textsc{Majority}\xspace}
\newcommand{\binomial}{\textsc{Beta}\xspace}
\newcommand{\entropy}{\textsc{Entropy}\xspace}
\newcommand{\dirichlet}{\textsc{Dirichlet}\xspace}
\newcommand{\Crp}{\textsc{Chinese Restaurant Process}\xspace}
\newcommand{\crp}{\textsc{crp}\xspace}

\newcommand{\codet}{CodeT\xspace}

\newcommand{\squishlist}{
  \begin{list}{$\bullet$}
    { \setlength{\itemsep}{0pt}      \setlength{\parsep}{3pt}
      \setlength{\topsep}{3pt}       \setlength{\partopsep}{0pt}
      \setlength{\leftmargin}{1.5em} \setlength{\labelwidth}{1em}
      \setlength{\labelsep}{0.5em} } }
\newcommand{\reallysquishlist}{
  \begin{list}{$\bullet$}
    { \setlength{\itemsep}{0pt}    \setlength{\parsep}{0pt}
      \setlength{\topsep}{0pt}     \setlength{\partopsep}{0pt}
      \setlength{\leftmargin}{0.2em} \setlength{\labelwidth}{0.2em}
      \setlength{\labelsep}{0.2em} } }

 \newcommand{\squishend}{
     \end{list} 
 }

 \newcommand{\symbolsecref}[1]{($\S$~\ref{#1})}

\newcommand{\numdsets}{17\xspace}

\newcommand{\nummodelswords}{three\xspace}

\begin{document}
\maketitle

\begin{abstract}
A popular approach for improving the correctness of output from large language models (LLMs) is \self{} -- poll the LLM multiple times and output the most frequent solution. Existing \self{} techniques always generate a \textit{constant} number of samples per question, where a better approach will be to non-uniformly distribute the available budget based on the amount of agreement in the samples generated so far. In response, we introduce \ours, a cost-efficient, model-agnostic technique that \textit{dynamically} adjusts the number of samples per question using a lightweight stopping criterion. Our experiments over \numdsets reasoning and code generation datasets and three LLMs demonstrate that \ours{} reduces sample budget by up to 7.9 times with an average accuracy drop of less than 0.1\%.\footnote{Code and \llm outputs are available at \url{https://sample-step-by-step.info/}.} 

%


\end{abstract}




\begin{figure*}[h]
    \centering
    \includegraphics[width=\linewidth]{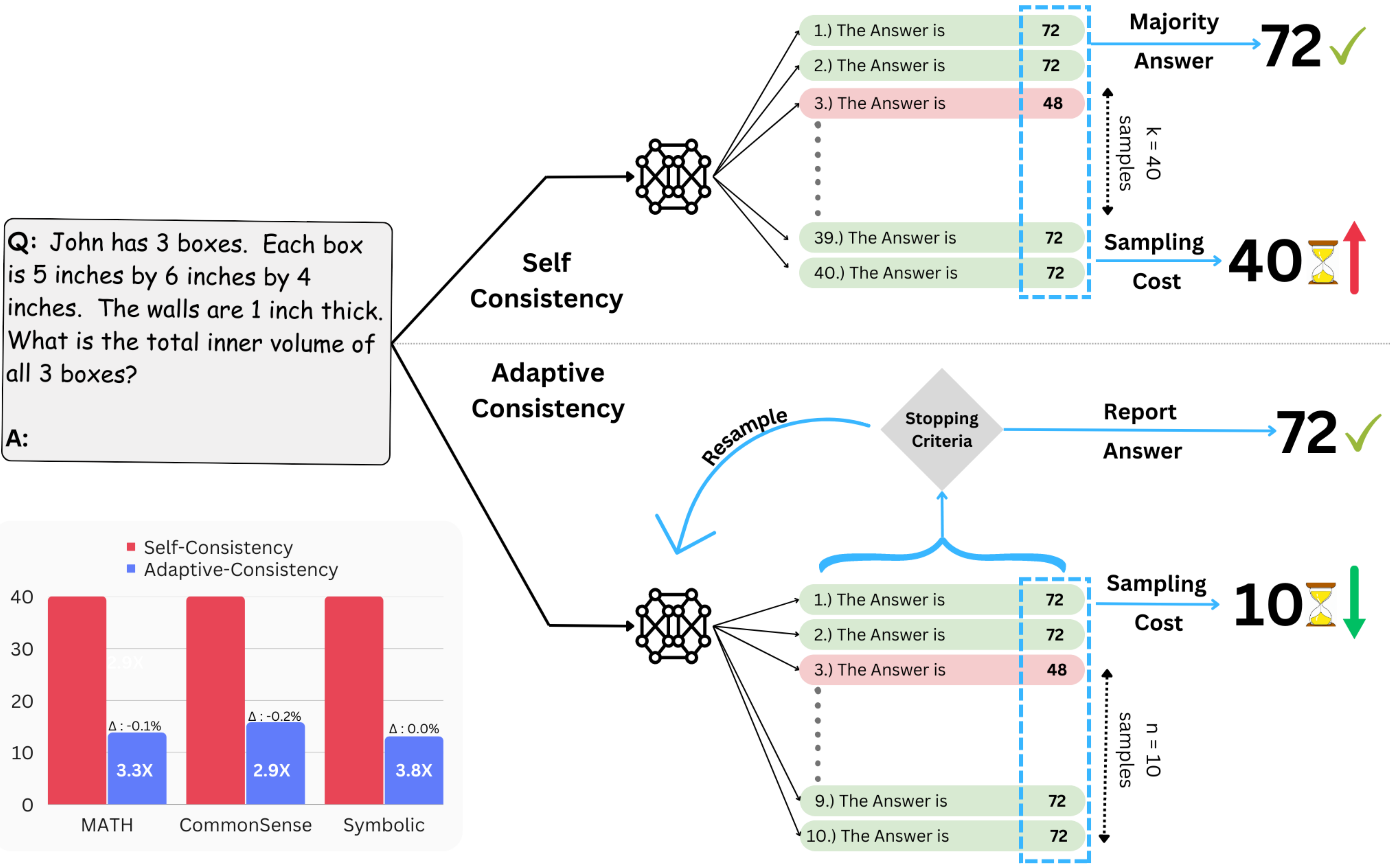}

    \caption{An overview of \textit{\ours}: \self samples a predetermined number of answers, whereas \ours iteratively samples until a lightweight \Stopfunc{}, decides to report the majority answer. 
    The figure demonstrates an example where \textit{\ours} reduces sampling costs by 4x, requiring only ten samples to report the majority answer.
    The bottom-left graph contrasts \textit{\ours} with \self across three reasoning categories, showing an average sample budget reduction of \boost{3.3} with a negligible $0.04\%$ drop in accuracy.
    }
    \label{fig:main_figure}
\end{figure*}

\section{Introduction}


The increasing adoption of large language models (LLMs) across various tasks, such as text generation and reasoning~\cite{Wei2022ChainOT,kojima2022large,wang2022rationale,mishra2022teachable}, mathematical reasoning~\citep{lewkowycz2022solving,Gao2022PALPL,Arora2023}, and code generation~\citep{Li2022CompetitionlevelCG,madaan2023learning}, has underscored the importance of improving the correctness of their outputs.
A popular method for achieving this goal is \textit{\self}~\cite{Wang2022SelfConsistencyIC}, a majority voting technique where multiple output samples are generated for a given input, and the final decision is based on the most frequently occurring output among the samples.

Current \self methods typically employ a fixed budget approach, wherein a predetermined number of samples (e.g., 40) are generated to make a decision. However, as LLMs continue to grow in size and complexity, the sampling time and computational costs associated with majority voting become increasingly challenging.
This challenge is particularly evident in high-stakes applications like competition-level code generation~\cite{Li2022CompetitionlevelCG}, where generating a large number of programs, sometimes up to a million, is essential for maximizing performance.

To address this challenge, we introduce \textit{\ours}, a cost-efficient, model-agnostic majority voting technique. \ours employs a lightweight stopping criterion that dynamically adjusts the number of samples ($n$) for each input, as opposed to using a fixed budget ($k$). The intuition is that if a clear majority is established with high confidence after sampling fewer than $k$ answers ($n < k$), there is no need to generate additional samples. 

\ours models the probability distribution over unique samples using a Dirichlet distribution, allowing us to quantify the confidence in the lead of the majority element over other elements. For instance, if the majority element has a count of 9 out of the first 10 samples, the likelihood of it remaining the majority element even after 40 samples is very high ($> 99\%$). This allows \ours to stop sampling at this point, reducing the cost by 30 samples, while \self would continue to sample all 40 answers. As an inference-time technique requiring no additional training, \ours provides a convenient off-the-shelf option for all pre-trained language models, offering the flexibility to balance computational cost and performance.

We evaluate \ours on \numdsets diverse tasks and three LLMs of different scales (\vicuna, \codex and \chatgpt).
Our experimental results show that \ours outperforms \self regarding cost efficiency while maintaining comparable output quality.
On \codex, \ours reduces the number of samples required by a factor of \boost{3.4}, with no average drop in accuracy.
On \vicuna, it requires sampling \boost{1.9} fewer samples, with almost no drop in accuracy.
Similarly, on \chatgpt, it samples \boost{4.4} fewer samples, with less than 0.2\% drop in accuracy. In summary, our contributions are:

\squishlist
    \item We propose \ours, a cost-efficient sampling technique for large language models that dynamically adjusts the number of samples using a lightweight stopping criterion based on the stability of the majority element.
    \item We conduct extensive experiments using three different \llms on a diverse set of 17 datasets. These datasets encompass a wide range of tasks, including \mathematical, \commonsense, \symbolic reasoning, and \coding tasks. \ours consistently and significantly outperforms fixed-budget methods like \self, requiring an average of \boost{3.3} fewer samples with less than 0.1\% drop in accuracy across all datasets and models.
    \item Our analysis reveals that for a fixed sampling cost, \ours consistently achieves better accuracy than \self across all datasets (upto 5\% absolute points). Additionally, we  experiment with various \stopfunc{}s and show the efficacy of \ours in terms of speed and accuracy.
\squishend

\begin{listing}
\begin{minipage}{\columnwidth}
\begin{mycode}
def adaptive_consistency(max_gens, 
                           stop_criterion):
    observations = []
    for k in range(1, max_gens):
        observations.append(sample_from_llm())
        if stop_criterion(observations):
            break
    return majority(observations)

def stop_criterion(observations, threshold):
    # Implement your stopping criterion
\end{mycode}
\end{minipage}
\begin{minipage}{\columnwidth}
\begin{mycode}
def self_consistency(max_gens):
    observations = []
    for k in range(1, max_gens):
        observations.append(sample_from_llm())
    return majority(observations)
\end{mycode}
\end{minipage}
\caption{Comparison of \ours (top) and \self (bottom). \self always generates a fixed number of samples. In contrast, \ours uses a lightweight stopping criterion, allowing it to adaptively halt the sampling process, which can lead to improved efficiency and performance.}
\label{lst:comparison}
\end{listing}



\section{Background}

\paragraph{In-Context Few-Shot Prompting}
In-context few-shot prompting is a technique employed by large language models (LLMs) to learn and generalize from a limited number of examples provided within the input of a given task. The model can quickly adapt to novel tasks without fine-tuning or additional training by conditioning the model on a few examples. Specifically, a prompt $p$ is constructed by concatenating multiple input-answer example pairs $<x_i, a_i>$.
The prompt is then prepended to the test input $x_{test}$, and the model generates the corresponding answer $a_{test}$.


\paragraph{Self-Consistency} 
\citet{Wang2022SelfConsistencyIC} proposed \self which improved performance 
by sampling multiple diverse reasoning chains and aggregating their outputs using a simple majority voting mechanism. However, higher accuracy is achieved with an increased computational cost, since the LLM must be prompted multiple times for the same question.

\section{\adaptive{}}

\self generates a predetermined number of answers ($k$) from the language model~(\llm) before returning the majority answer.
In contrast, the \ours method takes an incremental approach to sampling outputs from the language model. 
After generating each sample, \ours employs a lightweight \textit{\stopfunc{}} to determine whether it should 1.) generate an additional sample from \llm or 2.) cease sampling and report the current majority answer. 
This flexible strategy enables \ours to dynamically adjust the number of samples generated so far ($n$) for each input. As our experiments demonstrate, $n$ is typically less than $k$~(on average, \boost{3.3} and up to \boost{7.9} less in some cases), allowing \ours to offer greater cost-efficiency compared to the fixed budget approach employed by \self.

\ours  differs from \self only in terms of the \stopfunc{} (Listing~\ref{lst:comparison}). The design of the \stopfunc{} is crucial to our method, as it aims to minimize the average number of samples generated from the \llm while maximizing accuracy. 
The simplicity of our algorithm allows for the use of various \stopfunc{} interchangeably, each with its own advantages and disadvantages. We expand on a particular choice of stopping function next.

\paragraph{Dirichlet \Stopfunc{}}

Let $n$ be the number of samples generated from \llm so far, with $m$ unique samples.
Let $v = [v_1, v_2, \ldots, v_m]$ be the counts of each element, and $p_i = \frac{v_i}{n}$ be the normalized count.
For instance, if $n=10$, and $m=3$ (10 samples generated, with 3 unique elements), if $v = [8, 1, 1]$, then we can be more confident that $v_1$ is the answer.
On the other hand, if $v = [4, 4, 2]$, then more samples need to be generated.
Our goal is to formalize and quantify this intuition.

By convention, let $p_1 = \max(p_i)$. 
We want to assess the \textit{stability} of $p_1$ as the majority element.\footnote{Note that we overload the notation to use $p_1$ to represent both the majority element and its probability (usage clear from context).}
Specifically, we want to ask the following question: what is the probability that $p_1$ will be the majority element if we repeat the process of generating $n$ samples again?
Intuitively, if this probability is higher than some predetermined threshold $C_{thresh}$, then we can be more confident in our decision to stop sampling and return $p_1$ as the majority element:
\begin{equation*}
    P ( p_1 > \max_{i=2}^{m} p_i \mid  v) > C_{thresh}
 \label{eq:probeq}
 \end{equation*}

To answer this question, we establish a connection with the Dirichlet distribution. 
Specifically, we note that the counts $v$ parameterize a Dirichlet distribution, $\text{Dir}(V)$.\footnote{Dirichlet is a distribution over multinomials parameterized by counts $V$; each draw from Dirichlet is a multinomial distribution $p$. See Details in Appendix~\ref{app:dirichletproof}}
This connection allows us to explore the behavior of the sampling process by drawing more samples from $\text{Dir}(V)$ and observing the stability of $p_1$ as the majority element.
To compute the probability of $p_1$ being the majority element, we can integrate the joint probability density function of the Dirichlet distribution over the appropriate region of the probability simplex. The integral can be expressed as follows:

\begin{equation}
\begin{aligned}
P&(p_1 > \max_{i=2}^{m} p_i \mid V) \\
&= \int_{0}^{1} \int_{\mathcal{S}(p_1')} f(p_1', p_2, \ldots, p_m \mid V)  .\\
&\qquad\qquad\qquad\qquad dp_2 \cdots dp_m dp_1', \\
\text{where}\\ \; \mathcal{S}(p_1') &= \{ (p_2, \ldots, p_m) \mid p_1' > \max_{i=2}^{m} p_i, \\
& \qquad\qquad\qquad\qquad\qquad \sum_{i=2}^{m} p_i = 1 - p_1' \}.
\end{aligned}
\label{eq:dirichprob}
\end{equation}

In Equation~\ref{eq:dirichprob}, $f(p_1', p_2, ..., p_m | V)$ represents the joint probability density function of the Dirichlet distribution conditioned on the counts $V$.
The bounds on the integral for $p_1'$ range from 0 to 1. The probability simplex $\mathcal{S}(p_1')$ is defined for each $p_1'$ value, such that $p_1' > \max_{i=2}^{m} p_i$, and the remaining $p_i$ values sum to $1 - p_1'$. This constraint ensures that we are considering all possible values of $p_1'$ that would maintain its majority status. 
Here we assume, that the number of possible unique answers ($m$) is known, based on the current set of observations ($V$). 
In Analysis (\symbolsecref{analy:stopfuncs}, we further evaluate a \Crp (\crp) \stopfunc, which relaxes this assumption by not requiring the number of possible unique answers ($m$) to be known in advance.

\paragraph{Beta \Stopfunc{}}

Since the number of unique answers in the observation set can be large, Equation~\eqref{eq:dirichprob} is computationally expensive to solve.
As an approximation, we observe that establishing the majority of $p_1$ over the next largest probability, $p_2$, is sufficient for our purpose. 

In this setting, the probability in Equation~\eqref{eq:probeq} simplifies to a Beta distribution with parameters $(v_1 + 1, v_2 + 1)$, and Equation~\eqref{eq:dirichprob} is replaced by Equation~\eqref{eq:betaeq}.
This approximation, which assumes a non-informative prior of $\binomial{}(1, 1)$, allows us to efficiently compute the confidence in $p_1$ being the majority, enabling early stopping decisions without incurring substantial computational overhead.
\begin{equation}
    \int_{0}^{0.5} p_2^{v2} \cdot (1 - p_2)^{v_1} dp_{2}
\label{eq:betaeq}
\end{equation}

Empirically, we show the performance to be similar to Dirichlet \stopfunc{} but significantly faster (See Section~\ref{analy:stopfuncs}).
Throughout experiments, we refer to this Beta \Stopfunc{} as \ours.





\paragraph{Code-Generation}

We now turn our attention to \coding tasks, which involve generating programs that can correctly pass multiple test cases. More details on test case generation can be found in Appendix~\ref{app:coding_test}.

The configuration of code generation tasks significantly impacts the \self measurement since different programs might yield varying outputs for a given set of test cases.
This variation can cause simple majority voting schemes to be ineffective in evaluating stability. To address this, we explore two distinct methods for aggregating answers across multiple test cases.

In the first method, inspired by the approach used in AlphaCode~\cite{Li2022CompetitionlevelCG}, we concatenate the outputs for all test cases into a single vector with $t$ elements and apply \self across the entire vector. This implies that two programs are considered identical only if their outputs for all $t$ test cases match exactly. However, this simple setup may overestimate the output variance, as different programs can produce distinct outputs for the set of test cases.

To overcome the limitations of the simple setup, we propose an alternative method that treats test inputs as independent entities and applies \ours to each test case separately:

\begin{equation}
\sqrt[t]{\prod_{j=1}^{t} P(p_1^j > \max_{i=2}^{m} p_i^j \mid V)} \\
\label{eq:programprob}
\end{equation}

In this equation, $P$ is computed using Equation~\ref{eq:dirichprob}. The \ours method terminates the sampling process when the normalized probability---expressed as the geometric mean of $P$ across all $t$ test cases---exceeds a predefined threshold (e.g., 0.95).





\section{Experiments}


We evaluate \ours using \numdsets diverse benchmark datasets and three different language models.
We use prompts by program-aided language models, \pal{},~\cite{Gao2022PALPL}, \self{}~\cite{Wang2022SelfConsistencyIC} and \codet~\cite{Chen2022CodeTCG}.

\paragraph{Datasets}
\label{sec:datasets}

We evaluate our method on a diverse set of reasoning and coding benchmarks, encompassing 17 datasets across 4 distinct categories: \textbf{1. Mathematical Reasoning:} We use \gsm{}~\cite{Cobbe2021TrainingVT}, \svamp{}~\cite{Patel2021AreNM}, and \asdiv{}~\cite{miao-etal-2020-diverse} which assess the mathematical reasoning capabilities of the \llms. \textbf{2. \commonsense Reasoning Tasks}: We evaluate on 5 datasets: \strategyqa~\cite{Geva2021DidAU}, \dateunderstanding, \snarks, \ruinnames, \salienttranslation that measures different capabilites of \llms such as multi-hop reasoning and emotional understanding. \textbf{3. \symbolic Reasoning Tasks:} We further examine performance on 5 diverse \symbolic reasoning tasks: \trackingthree, \logicalthree, \booleanexpressions, \disambiguationqa, \penguins. \textbf{4. \coding Tasks} We also evaluate our method on coding tasks, which require to generate a working code given a textual problem description. We evaluate on 4 datasets of varying difficulty: \humaneval~\cite{Chen2021EvaluatingLL}, \mbpp{}~\cite{Austin2021ProgramSW}, \apps~\cite{Hendrycks2021MeasuringCC} and \codecontests~\cite{Li2022CompetitionlevelCG}. We refer readers to Appendix~\ref{sec:app_datasets} for more details.

\paragraph{Models}

We evaluate our method on three different language models:  \textbf{1. \chatgpt:}\footnote{\url{https://openai.com/blog/chatgpt}} An RLHF-finetuned GPT-3 based model (unreleased number of parameters). 
\textbf{2. \vicuna{}:}~\cite{vicuna2023} an open-source transformer model fine-tuned on instruction-following dataset~\cite{alpaca} from the base Llama series~\cite{Touvron2023LLaMAOA}.
\textbf{3. \codex{}:} A GPT-3-based publicly available model~\cite{brown2020language} which is a part of the Codex series~\cite{Chen2021EvaluatingLL} and has 175 billion parameters.\footnote{We have access to Codex models through OpenAI's researcher access program. Note that we only need access to the model outputs for this work, and we have released all outputs in the accompanying repository for reproducibility.} 

\begin{table*}
\small
\centering
\setlength\tabcolsep{2.5pt}
\begin{tabular}{lcccccc}
\toprule
& \multicolumn{2}{c}{Accuracy} & \multicolumn{2}{c}{Num. Generations} & \multicolumn{2}{c}{$\Delta$} \\
\cmidrule(lr){2-3} \cmidrule(lr){4-5} \cmidrule(lr){6-7}
Category & \makecell{\self} & \makecell{\ours} & \makecell{\self} & \makecell{\ours} & Num. Gen.  & Acc. \\
\midrule
\mathematical & \textbf{73.2} & 73.1 & 40 & \textbf{13.8} & 3.3$\times$ & -0.1 \\
\commonsense & \textbf{66.0} & 65.8 & 40 & \textbf{15.8} & 2.9$\times$ & -0.2 \\
\symbolic Reasoning  & 72.8 & 72.8 & 40 & \textbf{13.1} & 3.8$\times$ & +0.0 \\
\coding & 35.2 & \textbf{35.6} & 312.5 & \textbf{173.6} & 2.4$\times$ & +0.4 \\
\bottomrule
\end{tabular}

\vspace{0.4cm}

\begin{tabular}{lcccccc}
\toprule
& \multicolumn{2}{c}{Accuracy} & \multicolumn{2}{c}{Num. Generations} & \multicolumn{2}{c}{$\Delta$} \\
\cmidrule(lr){2-3} \cmidrule(lr){4-5} \cmidrule(lr){6-7}
Model & \makecell{\self} & \makecell{\ours} & \makecell{\self} & \makecell{\ours} & Num. Gen.  & Acc. \\
\midrule
\chatgpt & \textbf{76.4} & 76.2 & 40 & \textbf{10.0} & 4.4$\times$ & -0.2 \\
\vicuna & 54.0 & \textbf{54.1} & 40 & \textbf{21.7} & 1.9$\times$ & +0.0 \\
\codex & 69.7 & \textbf{69.8} & 104.1 & \textbf{49.4} & 3.4$\times$ & +0.0 \\
\bottomrule
\end{tabular}

\caption{\textbf{Main results:} \ours achieves a significant reduction in the number of generations, with a negligible impact on accuracy. The $\Delta$ columns display reductions in generations (Num. Gen.) and accuracy (Acc.) between \self and \ours. Detailed results are in Table~\ref{tab:fulltableseed}.}
\label{tab:mainresults}
\end{table*}

\paragraph{Prompting and Sampling}

We use similar prompts as in \pal{}~\cite{Gao2022PALPL} and \Cot{}~\cite{Wei2022ChainOT}. Specifically, for mathematical reasoning and \dateunderstanding tasks, we use  prompts from \pal{}. For other commonsense and \symbolic reasoning tasks, we use \CoT{}~\cite{Wei2022ChainOT}.

For sampling, we follow the scheme suggested in  ~\citet{Wang2022SelfConsistencyIC}. Specifically, we use a temperature of 0.7 for sampling and limit the number of generations to a maximum of 40. 
For coding tasks, we follow the exact procedure as used in CodeT~\cite{Chen2022CodeTCG}, with 50 samples for \apps, 100 samples for \humaneval and \mbpp and 1000 samples in \codecontests.



\paragraph{Hyperparameters}
\label{sec:hyperparameters}

The only hyperparameters in \ours{} are those related to parameters in \stopfunc{} ($C_{thresh}$). We use a high $C_{thresh} = 0.95$ for \ours{}.
By using a high threshold, we aim to maintain high accuracy and prevent the algorithm from stopping too early. For other \Stopfunc{}, we tune parameters on the training set of \gsm{}, and use the same thresholds across all the datasets. 
The impact of the chosen threshold on the performance of our method is further analyzed in the analysis section~\symbolsecref{anlys:sampbudgets}.


\paragraph{Baselines}

We compare our method against \self, which is the current state-of-the-art method.
Further, in Section~\ref{analy:stopfuncs}, we evaluate \ours against different \stopfunc{}, such as \random stopping and \majority (stopping at majority), \entropy, \dirichlet and \crp.

\paragraph{Evaluation Metrics}

We evaluate the performance of our method and the baselines using two metrics: average generations sampled from the LLMs, and overall reasoning accuracy. Our results show that \ours achieves similar performance to \self{} while often reducing sample budget considerably.

\subsection{Results}
\label{sec:results}


Table~\ref{tab:mainresults} presents the main results, and is divided into two parts showing results across different task categories~(top sub-table) and on various language models~(bottom sub-table). We focus on the potential tradeoff between efficiency and accuracy.

\paragraph{Results Across Task Categories} Our experimental results demonstrate the significant efficiency gains achieved by \ours across different task categories -- \boost{3.3} times fewer samples in mathematical tasks with a 0.1\% accuracy drop, \boost{2.9} times fewer samples in commonsense tasks with a 0.2\% accuracy drop, \boost{3.8} times fewer samples in symbolic reasoning tasks maintaining accuracy, and \boost{2.4} times fewer samples in coding tasks while improving accuracy by 0.4\%. These findings confirm the effectiveness of \ours in identifying the majority element early, highlighting its potential across various applications, including  reasoning and coding.

\paragraph{Results Across Language Models} Examining the results across different language models, we find that \ours is model-agnostic, and consistently reduces the number of generations with minimal to no impact on accuracy.
\ours consistently reduces the number of generations required, with reductions of \boost{4.4} for \chatgpt, \boost{1.9} for \vicuna, and \boost{3.4} for \codex, highlighting its cost-effective nature and adaptability to different scales of models. Moreover, the minimal accuracy differences and slight improvements showcase the practical utility of \ours, emphasizing its diverse applicability and model-agnostic characteristics.

\section{Analysis}

\subsection{Effect of Confidence Threshold in \ours}
\label{anlys:sampbudgets}

The confidence threshold, $C_{thresh}$, is a crucial hyperparameter for \ours, as it determines when to stop sampling based on the desired level of confidence in the majority element. 
While we set the threshold to a stringent value of 0.95 for all experiments, in this section, we analyze the impact of varying $C_{thresh}$ from 0.5 to 1 to understand the trade-offs between model accuracy and cost-efficiency.

\begin{figure}[t]
  \centering
  \includegraphics[width=0.99\linewidth]{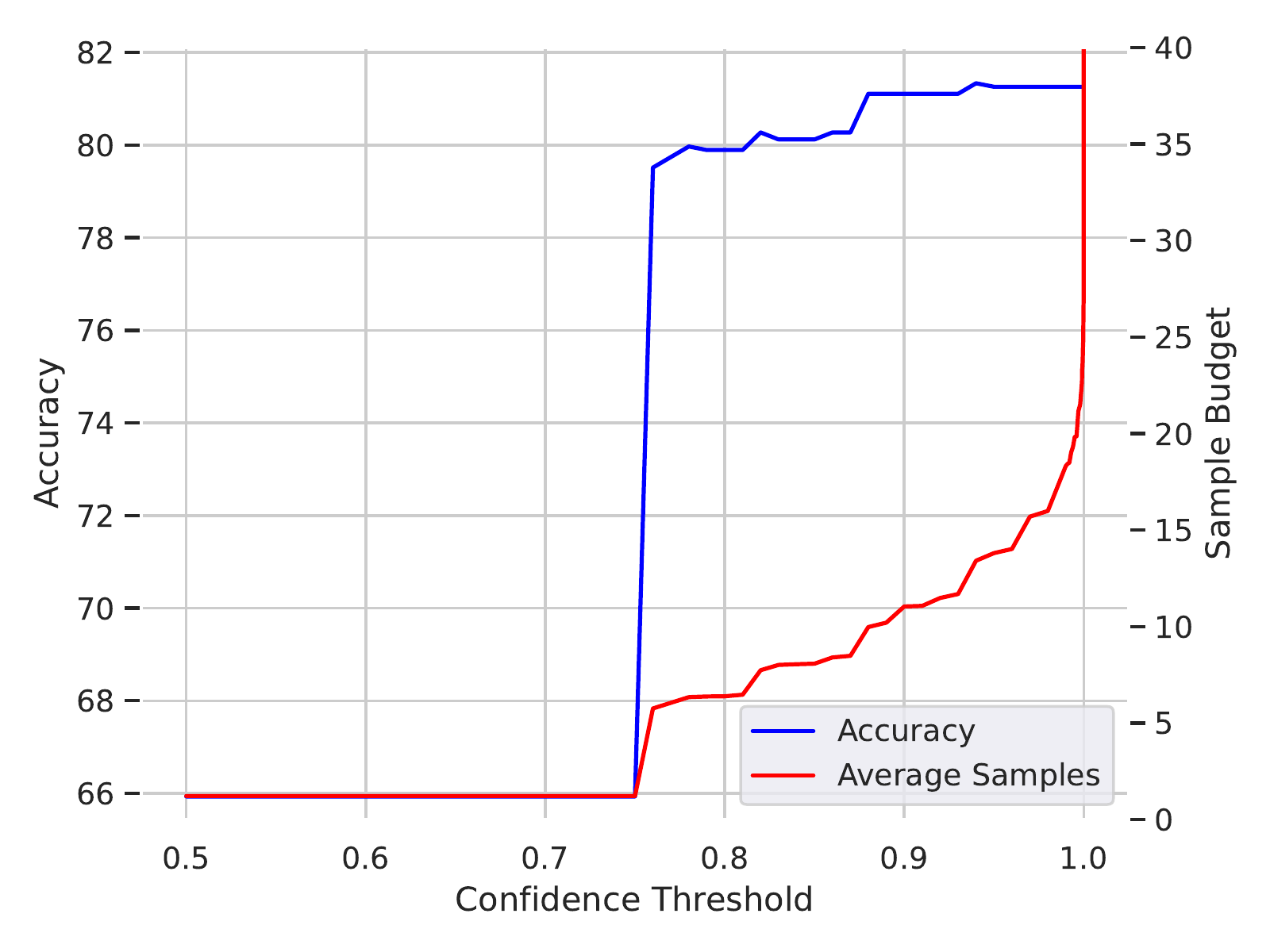}
  \caption{Impact of Confidence Threshold ($C_{thresh}$) on \ours for \gsm: As $C_{thresh}$ varies, the accuracy of \ours increases gradually, eventually plateauing. Initially, the average number of generations also increases gradually but then sharply climbs, reflecting the accuracy-confidence trade-off.}
  \label{fig:cthreshanalysis}
\end{figure}

In Figure \ref{fig:cthreshanalysis}, we present a visualization that examines the relationship between the confidence threshold, $C_{thresh}$, and the performance of adaptive consistency in terms of both accuracy and cost-efficiency. The x-axis represents the confidence threshold, varying from 0.5 to 1. The left y-axis displays the model's accuracy, while the right y-axis shows the average number of samples drawn.

The plot (for GSM-8K) shows the expected behavior of two curves: the blue curve (accuracy) increases gradually and then plateaus, while the red curve (average number of samples) initially increases linearly and then climbs more steeply. The plateau in accuracy signifies that the model has reached its maximum achievable accuracy, and further sampling will not improve it much. Meanwhile, the red curve's climbing rate indicates that the model requires more samples to meet an increasingly stringent confidence threshold for stopping, highlighting the trade-off between accuracy and cost efficiency. We refer readers to Appendix~\ref{app:confthresheffect} for more results.


\subsection{\ours vs. \self For Equal Average Sample Costs}
\label{anlys:cost_quality}

Section~\ref{sec:results} previously demonstrated that \ours achieves comparable performance to \self using fewer samples. \emnlpcr{In this section, our primary objective is to compare the performance of \ours{} to \self{} across various sampling budgets.} For each fixed sampling budget $k$, we contrast the performances of \ours and \self, where \self distributes sample budget uniformly to each question, \ours uses nonuniform allocation\emnlpcr{, rather than consistently across all instances.}

\begin{figure}[t]
    \centering
    \includegraphics[width=0.96\linewidth]{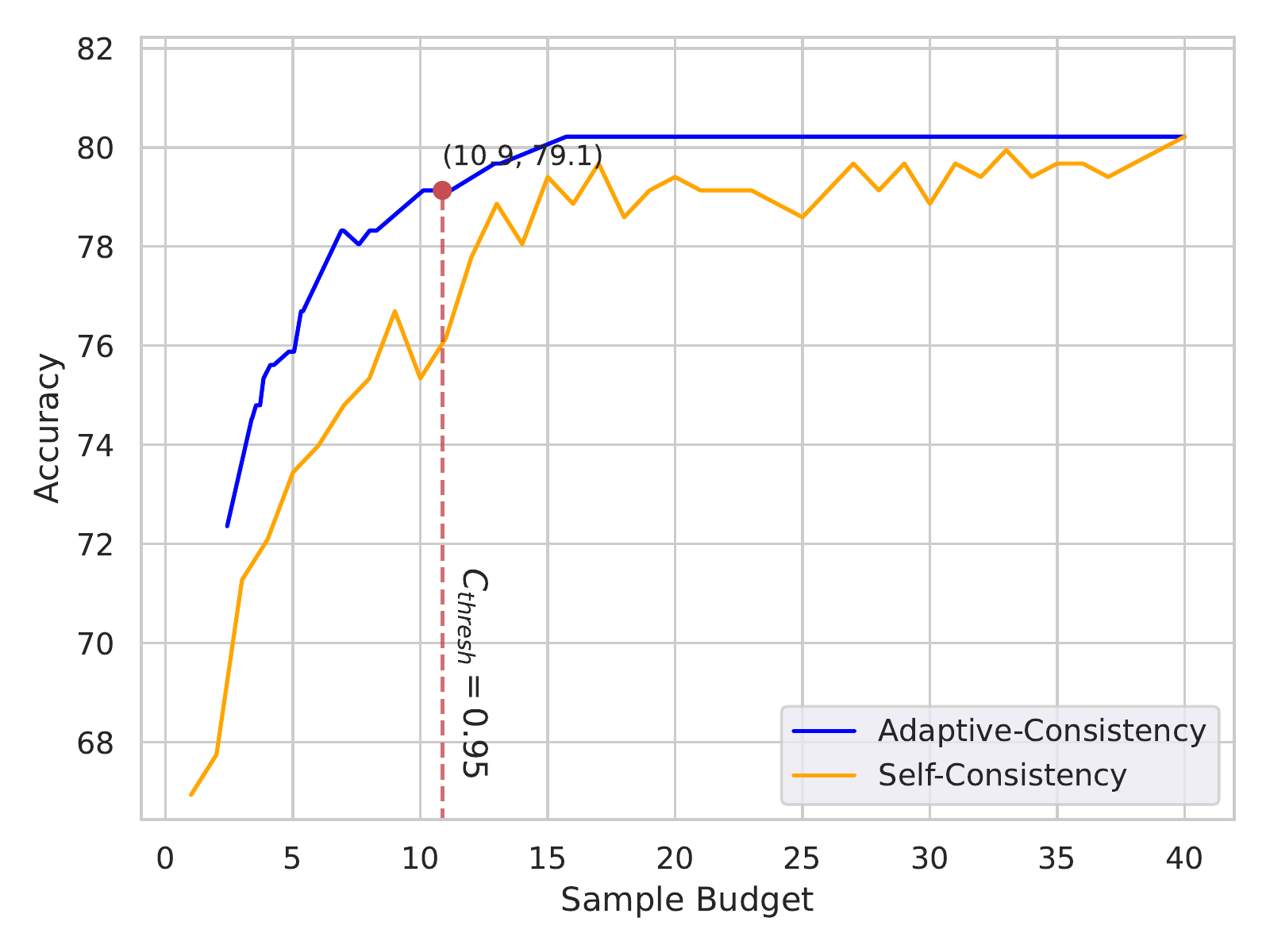}

    \caption{Comparison of \ours with \self on various average sampling costs on \emnlpcr{2 datasets}: \gsm and \dateunderstanding{}. \ours is able to consistently beat \self, especially when the sampling cost is low.}
    \label{fig:mathgraphs}
\end{figure}

\emnlpcr{We evaluate \ours using varying thresholds, with each threshold producing a distinct point (\#samples, performance) on the cost-quality curve. For every specific sample count (\#samples) generated by \ours, we subsequently run \self to obtain its corresponding performance. The relationship between the two methods across these data points is visualized in Figure~\ref{fig:mathgraphs}} which provides a visual comparison of the performance of \ours and \self on \gsm. \ours outperforms \self in accuracy across all average sample costs. 
For example, when the average sample cost is 10, \ours achieves approximately 3\% higher accuracy on \gsm. Similar results hold on other datasets; see Appendix~\ref{app:cost_quality} for full results.

The success of \ours can be attributed to the fact that it varies the number of samples based on the complexity of the instance, using more samples where a clear consensus is hard to reach and fewer where answers are consistent. Consequently, \ours achieves improved overall performance when controlled for cost budget. 

\subsection{Evaluation of Different Stopping Functions}
\label{analy:stopfuncs}

\ours allows a flexible choice of \stopfunc, based on intended objective and requirements. Here, we evaluate six different functions: 1) \random{}: randomly stopping with a probability $p$, 2) \majority{}: stopping after the most common answer has a majority above a threshold, 3) \entropy{}: stopping after the entropy of answers is below a threshold, 4) \binomial: The main stopping criterion used in \ours, based on Equation~\eqref{eq:betaeq}, 5) \dirichlet: The stopping criterion, based on Equation~\eqref{eq:dirichprob}, 6) \Crp (\crp): Unlike \dirichlet, \crp makes no assumption on the number of possible unique answers. Based on the available observations, we first model the concentration parameter ($\alpha$), denoting the probability of getting a new answer, then perform Monte Carlo simulations to obtain stability of the current majority (see Appendix~\ref{app:crp} for more details).

\begin{table}
    \centering
    \resizebox{0.99\linewidth}{!}{%
    \begin{tabular}{cccccc}
        \toprule
        & $\binomial$ & $\entropy$ & $\dirichlet$ & $\crp$ \\
        \midrule
        Time (ms) & 0.03 & 0.03 & 101.3 & 94.6 \\
        \bottomrule
    \end{tabular}
    }
    \caption{\emnlpcr{Time comparison of different \stopfunc{}s, when evaluated on \gsm and \dateunderstanding datasets. All \stopfunc{}s are significantly faster than \llm inference, with $\binomial$ being 4 orders of magnitude faster than other variants.}}
    \label{tab:timing_compare}
\end{table}

\begin{figure}[h]
  \centering
  \includegraphics[width=0.94\linewidth]{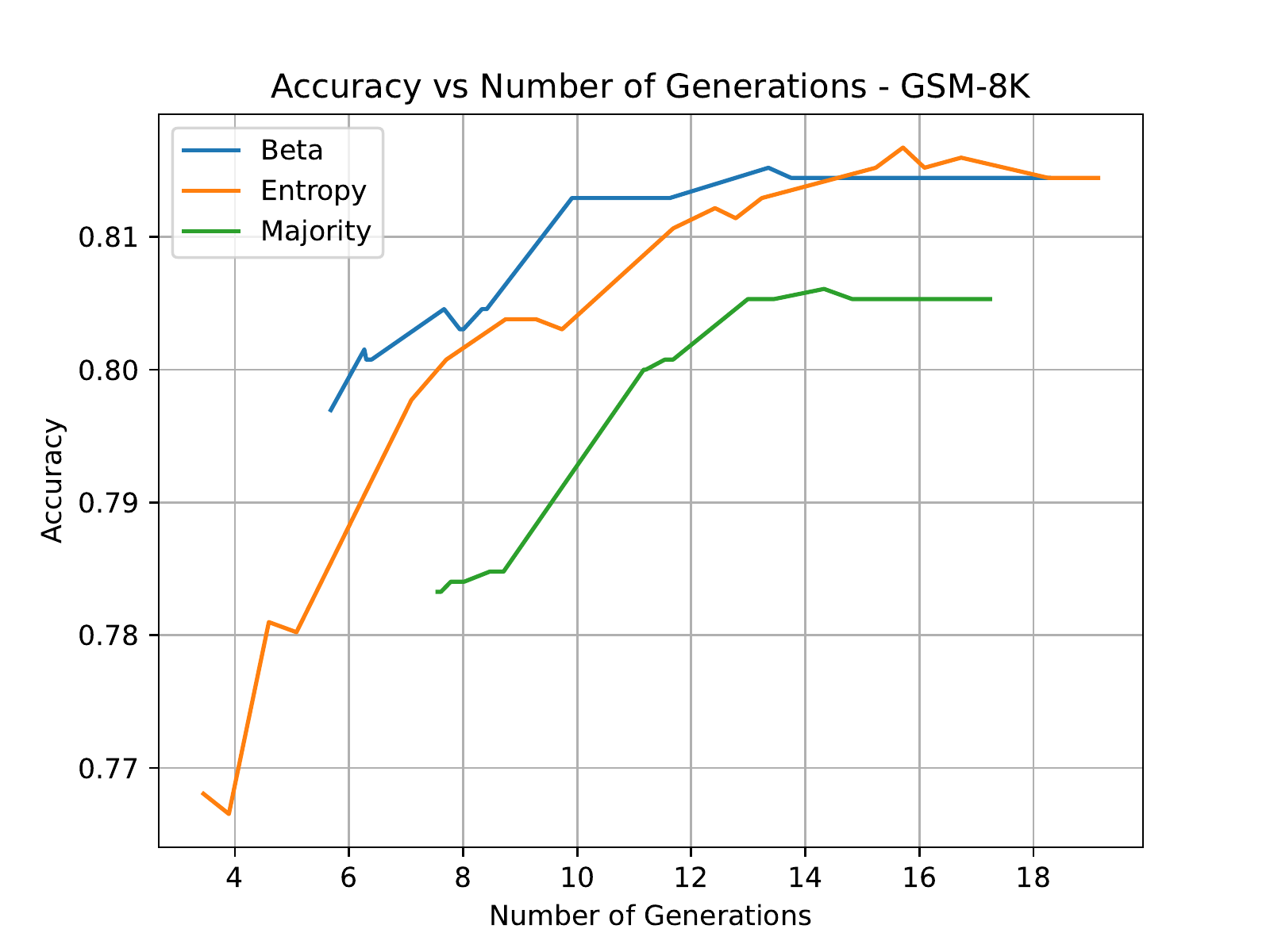}
  \caption{Comparison of \binomial, \entropy and \majority \stopfunc{}s. \binomial consistently beats \entropy and \majority  in terms of accuracy for the same sampling cost.}
  \label{anlys:betavsentropy}
\end{figure}

The parameters for all these methods are tuned, as discussed in Section~\ref{sec:hyperparameters}.
Figure~\ref{anlys:betavsentropy} compares \binomial to \entropy and \majority over a range of expected sampling costs. 
\binomial consistently achieves higher accuracy than both for the same sampling cost.
Further, we find \random to be the least effective method as expected, whereas \majority{} almost consistently underperforms both \binomial{} and \entropy{}.
While \dirichlet and \crp have a similar performance to \binomial, they are both about \emnlpcr{four} orders of magnitude slower than \binomial due to the expensive multivariate integral calculation. \emnlpcr{Nonetheless, despite being run on a single cpu core, even \dirichlet and \crp have negligible time and cost compared to \llm inference. The exact timings are presented in Table~\ref{tab:timing_compare}.}
The detailed results are presented in Appendix~\ref{app:stopfuncs}, Table~\ref{tab:stoptable}.

\emnlpcr{In summary, \ours is particularly effective in two scenarios: \textit{(i)} when a majority trend is evident early in the sampling process, such as in the \svamp dataset where it achieves comparable accuracy to Self-Consistency using fewer than 5 samples on average per input; and \textit{(ii)} for tasks with a limited set of potential answers, such as the \booleanexpressions dataset where \ours reduces the computational budget by 7.9 times without any loss in accuracy.
}

\section{Related Work}
\label{sec:related}

\paragraph{Crowdsourcing and Adaptive Consistency}
\textit{\ours} finds inspiration in techniques from crowdsourcing \cite{Lin2012CrowdsourcingCM, Dai2013POMDPbasedCO,weld-survey,Bragg2016Optimal}. Traditionally, crowdsourcing involves aggregating diverse human judgments, which presents challenges in managing resource allocation—knowing when to query additional contributors or stop based on the consistency of responses \cite{Doan2011CrowdsourcingSO,Quinn2011HumanCA}. Early research concentrated on probabilistic models estimating the 'true' answer and worker reliability \cite{Dawid1979MaximumLE,Whitehill2009WhoseVS}, later considering factors like worker expertise, task complexity, and answer quality \cite{Raykar2010LearningFC,Welinder2010TheMW}. However, rather than addressing these issues with multiple human contributors, \textit{\ours} is tailored specifically for LLMs, optimizing for computational efficiency and output accuracy.
\emnlpcr{In line with our vision, \cite{Parameswaran2023RevisitingPE} have recently proposed declarative prompt engineering, viewing LLMs like crowd workers and leveraging multiple prompting strategies.}

\paragraph{Architectures for adaptive computation}
A related body of work on adaptive computation aims to preempt computation based on intermediate representations~\citep{liu2020fastbert,zhou2020bert,schuster2021consistent,geng2021romebert, Xin2020DeeBERTDE}.
\citet{schuster2022confident} present \textsc{clam}, a language model that performs language generation adaptively.
\citet{Hou2020DynaBERTDB} propose Dynamic Bert, which can adapt the depth and width of the transformer to satisfy various computational constraints.
\citet{Xing2020EarlyEO} propose a dynamic deep neural network with an early-exit strategy embedded for enhancing the quality of compressed images.
Another direction of work focuses on pruning model weights or training sparse weights~\cite{Fan2019ReducingTD, Jayakumar2021TopKASTTA} to reduce training and inference time. 
In contrast to these methods, our approach completely obviates making any architectural modifications.

\paragraph{Inference-time adaptive computation}
These methods focus on adaptive computation at inference time without making architectural modifications to the models.
\citet{schwarzschild_can_2021,schwarzschild_datasets_2021} focus on three different generalization tasks.
They observe that increasing the number of test iterations (which corresponds to the network depth in their setting) helps the models in generalizing better to difficult problems.
\citet{madaan2022flowgen} leverage two different networks trained for the same task, a larger variant (slow) and a smaller variant (fast). The switch from fast to slow happens during inference, based on the complexity of generation at the current step.
\citet{Xue2023AdaptiveCW} train language models to adaptively read tokens from a tape bank for each input.
Different from these works, our focus is tasks where the multiple samples are drawn from a model~(vs. iteratively solving a task, which is a focus of these works).
\emnlpcr{Additionally, recent works such as \cite{Madaan2023AutoMixAM, Chen2023FrugalGPTHT} have propsed to adaptively selecting models of varying sizes based on verification signals derived from the output of the smaller model. Our methods, however, distinguish themselves by not necessitating the use of an additional verifier, and without the need of multiple models.}

\paragraph{Adaptive Sampling in Training and Active Learning}
\label{subsec:adaptive_sampling_active_learning}
Another line of work focuses on importance-based sampling of input instances during training~\citep{Bengio2008AdaptiveIS, Prabhu2019SamplingBI, Berger2017AnAS}. 
In contrast to the aforementioned methods, our approach centers on adaptively sampling multiple outputs per input instance during the inference phase, without soliciting additional labels. Our method is crafted to efficiently obtain reliable predictions from pretrained language models by adaptively sampling their outputs, distinguishing it from both adaptive sampling in training and active learning, which focus on the training phase.






\section{Conclusion and Future Work}

This paper presented \ours, a cost-efficient and model-agnostic technique for improving the correctness of output from large language models (\llms) using dynamic sampling. Our approach builds upon the \self{} method and introduces a lightweight stopping criterion that allows for adaptive sampling based on the amount of agreement in the samples drawn so far. \ours is effective across \numdsets datasets and \nummodelswords \llms, on both reasoning and coding tasks. It reduces the required sample budget by 2 to 4 times, while maintaining comparable accuracy, with an average drop of less than 0.1\%.

Our work opens up several avenues for future research. We may develop alternative stopping criteria, or combining multiple criteria could lead to even more efficient sampling techniques.
Moreover, in our current approach, the majority decision relies on using matches to determine the most common answer. However, this may not always capture the true majority, e.g., in generative tasks, where the output can have variations that do not affect the overall correctness or relevance of the answer. 
To foster further research and enable reproducibility, we have released the code and \llm outputs at \url{https://sample-step-by-step.info/}.

\section*{Acknowledgements}

We thank the anonymous reviewers for their useful comments and suggestions. 
Mausam is supported by grants from Microsoft, Google and Verisk, Wipro CoE on generative AI, Yardi School of AI travel funds, and the Jai Gupta
chair fellowship by IIT Delhi.  We thank the IIT Delhi HPC facility for its computational resources.
This work was also partially supported by the CSE Research Acceleration Fund of IIT Delhi
Aman is supported by a contract from the DARPA KAIROS program
under agreement number FA8750-19-2-0200. The U.S. Government is authorized to reproduce and
distribute reprints for Governmental purposes, notwithstanding any copyright notation thereon. The
views and conclusions contained herein are those of the authors and should not be interpreted as
necessarily representing the official policies or endorsements, either expressed or implied, of the U.S. Government.

\section*{Limitations}

Despite the promising results of our proposed \ours{} method, it \emnlpcr{bears} several limitations and \emnlpcr{scopes} for future improvement.

\squishlist

\item \textbf{Stopping criterion sensitivity:} \emnlpcr{The current stopping criterion, based on the majority element's stability in the sample set, may not always indicate sample agreement optimally. Instances may arise where the majority element lacks stability, yet the criterion triggers, potentially leading to suboptimal decisions. Future work could explore more robust or alternative stopping criteria.}

\item \textbf{Generalizability:} The effectiveness of our method may vary across tasks or models, despite testing on a diverse range of \numdsets datasets and \nummodelswords different \llms of contrastive scale. Notably, \ours is anticipated to fail where \self fails.

\item \textbf{Task-specific adaptations:} \emnlpcr{The task-agnostic nature of \ours{} might limit its performance on tasks that could benefit from task-specific adaptations. Specialized versions of \ours{} for specific tasks or domains could potentially enhance performance. 
We have initiated this by experimenting on \coding dataset, but extending \ours to other domains may not be as straightforward.}

\item \textbf{Reliance on the pretrained LLM:}  Our method depends on the pretrained LLM for generating multiple samples. Consequently, any limitations or biases in the LLM would persist in the \ours{}. 
Addressing these issues might require improvements in the LLM training process itself or the integration of external knowledge sources.

\squishend

\bibliography{custom}
\bibliographystyle{acl_natbib}

\clearpage

\appendix






\section{Experimental Setup}

\subsection{Hyperparameters}

The only hyperparameters in \ours, are those related to parameters in \stopfunc{}s ($C_{thresh}$). We use a high $C_{thresh} = 0.95$ for \ours{}.
By using a high threshold, we aim to maintain high accuracy and prevent the algorithm from stopping too early. For other \Stopfunc{}s, we tune our parameters on the training set of \gsm{}, and use the same thresholds across all the datasets. 
The impact of the chosen threshold on the performance of our method is further analyzed in the Analysis Section~\symbolsecref{anlys:sampbudgets}. We further evaluate all methods on a set of 3 seeds and report the table with standard deviation in Table~\ref{tab:fulltableseed}. We use only a single seed for \chatgpt because of the cost associated.

\subsection{Benchmarks}
\label{sec:app_datasets}

We evaluate our method on a diverse set of coding and reasoning benchmark datasets, encompassing \numdsets datasets across four distinct categories:

1. \textbf{MATHEMATICAL Reasoning:} To assess mathematical reasoning capabilities, we utilize the following datasets: \gsm{}~\cite{Cobbe2021TrainingVT}, \svamp{}~\cite{Patel2021AreNM}, and \asdiv{}~\cite{miao-etal-2020-diverse}. These datasets consist of grade-school-level algebra word problems necessitating arithmetic operations and problem-solving based on contextual information.

2. \textbf{\commonsense Reasoning Tasks:} We evaluate \ours{} on four \commonsense reasoning tasks. \textbf{1.) \strategyqa}~\cite{Geva2021DidAU} comprises questions that demand the model to infer a multi-hop strategy with reasoning steps implicitly embedded in the questions. \textbf{2.) \dateunderstanding} entails questions that require the model to deduce dates from natural language descriptions and perform arithmetic operations accordingly. \textbf{3.) \salienttranslation} is a salient translation error detection task that requires the model to identify the type of error in a translation. \textbf{4.) \snarks} and \textbf{5.) \ruinnames} both focus on emotional understanding tasks.

\begin{table}[t]
    \tiny
  \centering
  \resizebox{0.99\linewidth}{!}{
    \begin{tabular}{@{}|l|c|c|}
    \hline
    Dataset & $|N\_test|$ & Answer Format  \\

    \hline
\gsm & 1319 & Numerical \\
\hline
\asdiv & 2096 & Numerical \\
\hline
\svamp & 1000 & Numerical \\
\hline
\dateunderstanding & 369 & String \\
\hline
\trackingthree & 250 & MCQ \\
\hline
\logicalthree & 250 & MCQ \\
\hline
\strategyqa & 2279 & Binary \\
\hline
\booleanexpressions & 250 & Binary \\
\hline
\snarks & 250 & Binary \\
\hline
\ruinnames & 178 & MCQ \\
\hline
\salienttranslation & 250 & MCQ \\
\hline
\disambiguationqa & 250 & MCQ \\
\hline
\penguins & 146 & MCQ \\
\hline
\humaneval & 164 & Python Code \\
\hline
\mbpp & 427 & Python Code \\
\hline
\apps & 5000 & Coding \\
\hline
\codecontests & 165 & Competitive Coding \\
\hline
    
  \end{tabular}
  }
\caption{Dataset Statistics. We evaluate on \numdsets diverse reasoning datasets, having different difficulty, domains, answer types, sizes ($N\_test$). 
}
 \label{tab:datasetstats}
\end{table}

3. \textbf{\symbolic Reasoning Tasks:} We examine the performance of our method on six diverse \symbolic reasoning tasks. \textbf{1.) \trackingthree} is a tracking task that necessitates the model to infer the final state of a system, given its initial state and a sequence of modifications. \textbf{2.) \logicalthree} is a logical deduction task that demands the model to deduce the order of a sequence of objects based on a minimal set of conditions. \textbf{3.) \booleanexpressions} is a boolean expressions task that evaluates whether a language model has learned the rules of deductive reasoning, i.e., formal (zeroth-order) logic associated with the words "and," "or," "not," etc. \textbf{4.) \disambiguationqa} is a disambiguation task that necessitates the model to select the person to whom the pronoun refers. \textbf{5.) \penguins} describes a table of penguins and requires the model to answer questions about the penguins' attributes.

4. \textbf{\coding Tasks:} We further evaluate the performance of our method by conducting experiments on four diverse standard coding tasks. These tasks encompass a range of programming challenges, including both basic human-written and crowd-sourced Python tasks found in the \textbf{1.) \humaneval}~\cite{Chen2021EvaluatingLL} and \textbf{2.) \mbpp{}}~\cite{Austin2021ProgramSW} datasets, as well as more challenging competition-level coding tasks from the \textbf{3.) \apps}~\cite{Hendrycks2021MeasuringCC} and \textbf{4.) \codecontests}~\cite{Li2022CompetitionlevelCG} datasets.

\subsection{Tools and Framework}

For querying \chatgpt and \codex models~\cite{Chen2021EvaluatingLL}, we use the api library provided by OpenAI\footnote{API available at: \href{https://platform.openai.com/}{https://platform.openai.com/}}. 
We use the official code provided for running \vicuna model~\cite{vicuna2023}. 
We run inference on \vicuna models on single A100 gpus. 
For coding tasks, we use the outputs provided by \codet~\cite{Chen2022CodeTCG}, where models are zero-shot prompted with temperature=0.8, and top\_p = 0.95.
\stopfunc{} in \ours are fast to run, and we use a single-core machine. For numerical integration, we use the Scipy library in Python.

\subsection{Test-Case Generation}
\label{app:coding_test}

For \coding tasks, we generate test cases in a similar fashion to \codet~\cite{Chen2022CodeTCG}. Specifically, we prompt the model with function description and prompt for generation of assert statements. However, unlike \codet, we limit ourselves to only 10 test cases, which are generated in 1-2 prompts to LLM, thus adding neglible effect on the code generation itself.


Dataset Statistics are presented in Table~\ref{tab:datasetstats}.




\section{Results}

\begin{table*}[ht!]
  \centering
  \resizebox{0.99\linewidth}{!}{
    \begin{tabular}{@{}lc|cc|cc|cc|cc@{}}
    \toprule
      & & \multicolumn{2}{c|}{\textbf{\self}} & \multicolumn{2}{c|}{\textbf{\ours}} & \multicolumn{2}{c}{$\Delta$} \\
     \cmidrule(lr){3-4} \cmidrule(lr){5-6} \cmidrule(lr){7-8}
     & Model & \textbf{Avg. Gen.} & \textbf{{Accuracy}} & \textbf{Avg. Gen.} & \textbf{{Accuracy}} &\textbf{Gen. Reduc.} & \textbf{{Acc. Diff. $\uparrow$}} \\
    \midrule
\addlinespace
\multirow{3}{*}{\textbf{\humaneval}} & \codex & 100 & $61.4$ & $\textbf{23.6}$ & $\textbf{63.4}$ & \boost{$4.3$} & $+2.0$ \\
\addlinespace
 & \incoder & 100 & $19.5$ & $\textbf{51.2}$ & $\textbf{20.1}$ & \boost{$2.0$} & $+0.6$ \\
\addlinespace
 & \codegen & 100 & $34.1$ & $\textbf{54.7}$ & $\textbf{36.0}$ & \boost{$1.8$} & $+1.9$ \\
\addlinespace
\hline
\addlinespace
\multirow{3}{*}{\textbf{\mbpp}} & \codex & 100 & $\textbf{64.4}$ & $\textbf{36.3}$ & $63.9$ & \boost{$2.8$} & $-0.5$ \\
\addlinespace
 & \incoder & 100 & $30.7$ & $\textbf{53.8}$ & $\textbf{30.9}$ & \boost{$1.9$} & $+0.2$ \\
 \addlinespace
 & \codegen & 100 & $49.6$ & $\textbf{57.8}$ & $\textbf{50.4}$ & \boost{$1.7$} & $+0.8$ \\
\addlinespace
\hline
\addlinespace
\textbf{\apps} & \codex & 50 & $\textbf{11.9}$ & $\textbf{44.4}$ & $\textbf{11.9}$ & \boost{$1.1$} & $0.0$ \\
\addlinespace
\hline
\addlinespace
\textbf{\codecontests} & \codex & 1000 & $\textbf{3.0}$ & $\textbf{590.2}$ & $\textbf{3.0}$ & \boost{$1.6$} & $0.0$ \\
\addlinespace
\hline

\bottomrule
  \end{tabular}
  }
\caption{Comparison of \ours with \self on 4 diverse code generation datasets. The table presents the accuracy of \self, the average number of generations (Avg. Gen.) for \ours, and the accuracy of \ours. \self always draws a fixed number of samples. The $\Delta$ columns display the reduction in generations (Gen. Reduc.) and the difference in accuracy (Acc. Diff.) between \self and \ours. For \codecontests, \self uses 1000, \apps use 50, while \humaneval and \mbpp use 100 generations each. 
}
 \label{tab:fulltable_code}
\end{table*}

We present the complete results with standard deviation in Table~\ref{tab:fulltableseed}. 
For \coding tasks, results are presented in Table~\ref{tab:fulltable_code}


\begin{table*}[ht!]
  \centering
  \resizebox{0.99\linewidth}{!}{
    \begin{tabular}{@{}lc|c|cc|cc|cc@{}}
    \toprule
      & & \multicolumn{1}{c|} {\textbf{\self}} & \multicolumn{2}{c|}{\textbf{\ours}} & \multicolumn{2}{c}{$\Delta$} \\
     \cmidrule(lr){3-3} \cmidrule(lr){4-5} \cmidrule(lr){6-7}
     & & \textbf{{Accuracy}} & \textbf{Avg. Gen.} & \textbf{{Accuracy}} &\textbf{Gen. Reduc.} & \textbf{{Acc. Diff. $\uparrow$}} \\
    \midrule
\addlinespace
\multirow{3}{*}{\textbf{\gsm}} & \vicuna & $\textbf{31.6}$\tablestd{3.0} & $\textbf{26.8}$\tablestd{2.2} & $31.5$\tablestd{3.0} & \boost{$1.4$} & $-0.1$ \\
\addlinespace
 & \codex & $\textbf{81.1}$\tablestd{0.3} & $\textbf{13.8}$\tablestd{0.0} & $81.0$\tablestd{0.3} & \boost{$2.9$} & $-0.1$ \\
\addlinespace
 & \chatgpt & $\textbf{82.7}$ & $\textbf{9.2}$ & $\textbf{82.7}$ & \boost{$4.3$} & $0.0$ \\
 \addlinespace
\hline
\addlinespace
\multirow{3}{*}{\textbf{\svamp}} & \vicuna & $\textbf{63.0}$\tablestd{0.3} & $\textbf{18.8}$\tablestd{0.1} & $62.8$\tablestd{0.4} & \boost{$2.1$} & $-0.2$ \\
\addlinespace
 & \codex & $\textbf{85.1}$\tablestd{0.3} & $\textbf{9.5}$\tablestd{0.1} & $85.0$\tablestd{0.3} & \boost{$4.2$} & $-0.1$ \\
\addlinespace
 & \chatgpt & $\textbf{85.1}$ & $\textbf{9.5}$ & $85.0$ & \boost{$4.2$} & $-0.1$ \\
\addlinespace
\hline
\addlinespace
\multirow{3}{*}{\textbf{\asdiv}} & \vicuna & $\textbf{64.0}$\tablestd{0.3} & $\textbf{16.5}$\tablestd{0.2} & $\textbf{64.0}$\tablestd{0.3} & \boost{$2.4$} & $0.0$ \\
\addlinespace
 & \codex & $\textbf{83.2}$\tablestd{0.2} & $\textbf{10.0}$\tablestd{0.0} & $83.2$\tablestd{0.2} & \boost{$4.0$} & $0.0$ \\
\addlinespace
& \chatgpt & $\textbf{83.0}$ & $\textbf{10.0}$ & $\textbf{83.0}$ & \boost{$4.0$} & $0.0$ \\
\addlinespace
\hline
\addlinespace
\multirow{3}{*}{\textbf{\dateunderstanding}} & \vicuna & $59.8$\tablestd{0.3} & $\textbf{17.3}$\tablestd{0.3} & $\textbf{60.2}$\tablestd{0.4} & \boost{$2.3$} & $+0.4$ \\
\addlinespace
 & \codex & $\textbf{80.3}$\tablestd{0.1} & $\textbf{10.7}$\tablestd{0.3} & $79.5$\tablestd{0.3} & \boost{$3.7$} & $-0.8$ \\
\addlinespace
& \chatgpt & $\textbf{77.5}$ & $\textbf{9.1}$ & $77.0$ & \boost{$4.4$} & $-0.5$ \\
\addlinespace
\hline
\addlinespace
\multirow{3}{*}{\textbf{\trackingthree}} & \vicuna & $31.8$\tablestd{1.0} & $\textbf{20.3}$\tablestd{0.0} & $\textbf{32.0}$\tablestd{1.2} & \boost{$2.0$} & $+0.2$ \\
\addlinespace
 & \codex & $\textbf{77.2}$\tablestd{1.3} & $\textbf{9.7}$\tablestd{0.1} & $77.1$\tablestd{1.6} & \boost{$4.1$} & $-0.1$ \\
\addlinespace
 & \chatgpt & $85.2$ & $\textbf{6.2}$ & $\textbf{85.6}$ & \boost{$6.4$} & $+0.4$ \\
\addlinespace
\hline
\addlinespace
\multirow{3}{*}{\textbf{\logicalthree}} & \vicuna & $51.2$\tablestd{0.8} & $\textbf{18.1}$\tablestd{0.2} & $\textbf{51.4}$\tablestd{0.6} & \boost{$2.2$} & $+0.2$ \\
\addlinespace
 & \codex & $\textbf{89.4}$\tablestd{0.2} & $\textbf{8.5}$\tablestd{0.1} & $\textbf{89.4}$\tablestd{0.2} & \boost{$4.7$} & $0.0$ \\
\addlinespace
  & \chatgpt & $\textbf{86.8}$ & $\textbf{7.5}$ & $\textbf{86.8}$ & \boost{$5.3$} & $0.0$ \\
\addlinespace
\hline
\addlinespace
\multirow{3}{*}{\textbf{\strategyqa}} & \vicuna & $65.8$\tablestd{0.5} & $\textbf{16.3}$\tablestd{0.1} & $\textbf{65.8}$\tablestd{0.4} & \boost{$2.5$} & $0.0$ \\
\addlinespace
 & \codex & $\textbf{79.0}$\tablestd{0.2} & $\textbf{11.9}$\tablestd{0.2} & $78.8$\tablestd{0.1} & \boost{$3.4$} & $-0.2$ \\
\addlinespace
& \chatgpt & $\textbf{68.1}$ & $\textbf{11.8}$ & $67.9$ & \boost{$3.4$} & $-0.2$ \\
\addlinespace
\hline
\addlinespace
\multirow{3}{*}{\textbf{\booleanexpressions}} & \vicuna & $\textbf{79.2}$\tablestd{0.6} & $\textbf{16.2}$\tablestd{0.3} & $78.4$\tablestd{0.3} & \boost{$2.5$} & $-0.8$ \\
\addlinespace
 & \codex & $\textbf{94.5}$\tablestd{0.4} & $\textbf{6.6}$\tablestd{0.1} & $\textbf{94.5}$\tablestd{0.4} & \boost{$6.0$} & $0.0$ \\
\addlinespace
 & \chatgpt & $\textbf{93.2}$ & $\textbf{5.0}$ & $92.8$ & \boost{$7.9$} & $-0.4$ \\
\addlinespace
\hline
\addlinespace
\multirow{3}{*}{\textbf{\snarks}} & \vicuna & $73.2$\tablestd{1.0} & $\textbf{23.2}$\tablestd{0.7} & $\textbf{73.6}$\tablestd{0.8} & \boost{$1.7$} & $+0.4$ \\
\addlinespace
 & \codex & $\textbf{74.0}$\tablestd{1.0} & $\textbf{12.7}$\tablestd{0.4} & $\textbf{74.0}$\tablestd{1.5} & \boost{$3.1$} & $0.0$ \\
\addlinespace
& \chatgpt & $\textbf{65.7}$ & $\textbf{8.8}$ & $65.2$ & \boost{$4.5$} & $-0.6$ \\
\addlinespace
\hline
\addlinespace
\multirow{3}{*}{\textbf{\ruinnames}} & \vicuna & $\textbf{43.6}$\tablestd{2.1} & $\textbf{33.8}$\tablestd{0.6} & $\textbf{43.6}$\tablestd{2.1} & \boost{$1.2$} & $0.0$ \\
\addlinespace
 & \codex & $\textbf{78.0}$\tablestd{0.9} & $\textbf{17.2}$\tablestd{0.1} & $\textbf{78.0}$\tablestd{0.6} & \boost{$2.3$} & $0.0$ \\
\addlinespace
  & \chatgpt & $\textbf{74.8}$ & $\textbf{13.1}$ & $74.0$ & \boost{$3.1$} & $-0.8$ \\
\addlinespace
\hline
\addlinespace
\multirow{3}{*}{\textbf{\salienttranslation}} & \vicuna & $\textbf{28.9}$\tablestd{2.4} & $\textbf{28.7}$\tablestd{2.5} & $28.7$\tablestd{2.5} & \boost{$1.2$} & $-0.3$ \\
\addlinespace
 & \codex & $\textbf{64.3}$\tablestd{0.2} & $\textbf{11.8}$\tablestd{0.5} & $\textbf{64.3}$\tablestd{0.2} & \boost{$3.4$} & $0.0$ \\
\addlinespace
  & \chatgpt & $\textbf{56.8}$ & $\textbf{11.1}$ & $\textbf{56.8}$ & \boost{$3.6$} & $0.0$ \\
\addlinespace
\hline
\addlinespace
\multirow{3}{*}{\textbf{\disambiguationqa}} & \vicuna & $\textbf{63.7}$\tablestd{0.7} & $\textbf{22.8}$\tablestd{1.0} & $63.5$\tablestd{1.1} & \boost{$1.8$} & $-0.3$ \\
\addlinespace
 & \codex & $74.9$\tablestd{0.8} & $\textbf{13.5}$\tablestd{0.6} & $\textbf{75.1}$\tablestd{0.7} & \boost{$3.0$} & $+0.1$ \\
\addlinespace
 & \chatgpt & $\textbf{62.5}$ & $\textbf{13.9}$ & $\textbf{62.5}$ & \boost{$2.9$} & $0.0$ \\
\addlinespace
\hline
\addlinespace
\multirow{3}{*}{\textbf{\penguins}} & \vicuna & $46.8$\tablestd{1.8} & $\textbf{22.9}$\tablestd{0.7} & $\textbf{47.3}$\tablestd{1.9} & \boost{$1.7$} & $+0.5$ \\
\addlinespace
 & \codex & $83.8$\tablestd{0.9} & $\textbf{11.0}$\tablestd{0.4} & $\textbf{84.0}$\tablestd{0.6} & \boost{$3.6$} & $+0.2$ \\
\addlinespace
& \chatgpt & $\textbf{71.9}$ & $\textbf{14.2}$ & $\textbf{71.9}$ & \boost{$2.8$} & $0.0$ \\
\addlinespace
\hline
    \addlinespace
    \multicolumn{2}{l|}{\textbf{Average}} & \textbf{70.3}\tablestd{0.8} & \textbf{14.1}\tablestd{0.5} & 70.2\tablestd{0.8} & \textbf{\boost{3.2}} & \textbf{-0.07} \\
    \addlinespace

\bottomrule
  \end{tabular}
  }
\caption{Comparison of \ours with \self on \numdsets diverse coding \& reasoning datasets. \self always draws 40 samples. The table shows accuracy, average generations (Avg. Gen.). The $\Delta$ columns display reductions in generations (Gen. Reduc.) and accuracy (Acc. Diff.) between \self and \ours. \ours achieves a \boost{3.2} reduction in sample budget (\textit{Gen. Reduc.}) with minimal average accuracy drop of 0.07\% (\textit{Acc. Diff.}).}
 \label{tab:fulltableseed}
\end{table*}

\begin{table*}[h]
    \centering
      \resizebox{0.99\linewidth}{!}{
    \begin{tabular}{l|l|c|c|}
    \toprule
    \textbf{Dataset}             & \textbf{Model}                     & \textbf{P-Value (Accuracy)} & \textbf{P-Value (Num Gens)} \\
    \hline
    \gsm                 & \vicuna                & 0.5                & 0.0056             \\
    \hline
    \gsm                 & \codex   & 0.42               & 2.09E-06           \\
    \hline
    \svamp               & \vicuna                & 0.07               & 3.47E-06           \\
    \hline
    \svamp               & \codex   & 0.42               & 7.40E-06           \\
    \hline
    \asdiv               & \vicuna                & 1                  & 0.0005             \\
    \hline
    \asdiv               & \codex   & 1                  & 0.0023             \\
    \hline
    \dateunderstanding                & \vicuna                & 0.057              & 7.68E-05           \\
    \hline
    \dateunderstanding                & \codex   & 0.04               & 4.63E-05           \\
    \hline
    \trackingthree      & \vicuna                & 0.5                & 0.00002            \\
    \hline
    \trackingthree      & \codex   & 0.67               & 9.88E-06           \\
    \hline
    \logicalthree       & \vicuna                & 0.5                & 0.0007             \\
    \hline
    \logicalthree       & \codex   & -                  & 0.0016             \\
    \hline
    \strategyqa         & \vicuna                & 0.90               & 1.16E-05           \\
    \hline
    \strategyqa         & \codex   & 0.24               & 0.0005             \\
    \hline
    \booleanexpressions & \vicuna                & 0.32               & 8.52E-05           \\
    \hline
    \booleanexpressions & \codex   & -                  & 4.98E-06           \\
    \hline
    \snarks              & \vicuna                & 0.18               & 0.0007             \\
    \hline
    \snarks              & \codex   & 1                  & 0.0001             \\
    \hline
    \ruinnames          & \vicuna                & -                  & 0.0049             \\
    \hline
    \ruinnames          & \codex   & 1                  & 8.72E-06           \\
    \hline
    \salienttranslation & \vicuna                & 0.18               & 0.0211             \\
    \hline
    \salienttranslation & \codex   & 1                  & 0.0001             \\
    \hline
    \disambiguationqa      & \vicuna                & 0.53               & 0.0015             \\
    \hline
    \disambiguationqa      & \codex   & 0.42               & 0.0002             \\
    \hline
    \penguins            & \vicuna                & 0.18               & 0.0009             \\
    \hline
    \penguins            & \codex   & 0.42               & 7.79E-05           \\
    \hline
    \textbf{Average}             &                           & 0.503              & 0.0002             \\
    \bottomrule
    \end{tabular}}
    \caption{\emnlpcr{P-values using 2 sample t-test over 3 seeds on multiple datasets and models. The p-value for 'number of generations' is significantly less than 0.05 (average: 1.5e-3), confirming our method's efficiency, while the p-value for accuracy is much larger than 0.05 (average: 0.50), indicating that the slight accuracy difference is statistically insignificant.}}
    \label{tab:ptable}    
\end{table*}

\emnlpcr{Further in Table~\ref{tab:ptable} we show that improvements by \ours are statistically significant across all datasets. We perform 2 sample t-test on 3 random seeds. While p-value of number of generations is much less than 0.05 (average: 1.5e-3), indicating that our method is significantly more efficient, the p-value of accuracy is much larger than 0.05 (average: 0.50), indicating that the slight accuracy difference between baseline and our method is statistically insignificant.}
\section{Analysis}

\subsection{\ours vs. \self For Equal Average Sample Costs}
\label{app:cost_quality}

\begin{figure*}[h]
  \centering
  \subfigure[\gsm]{\includegraphics[width=0.32\linewidth]{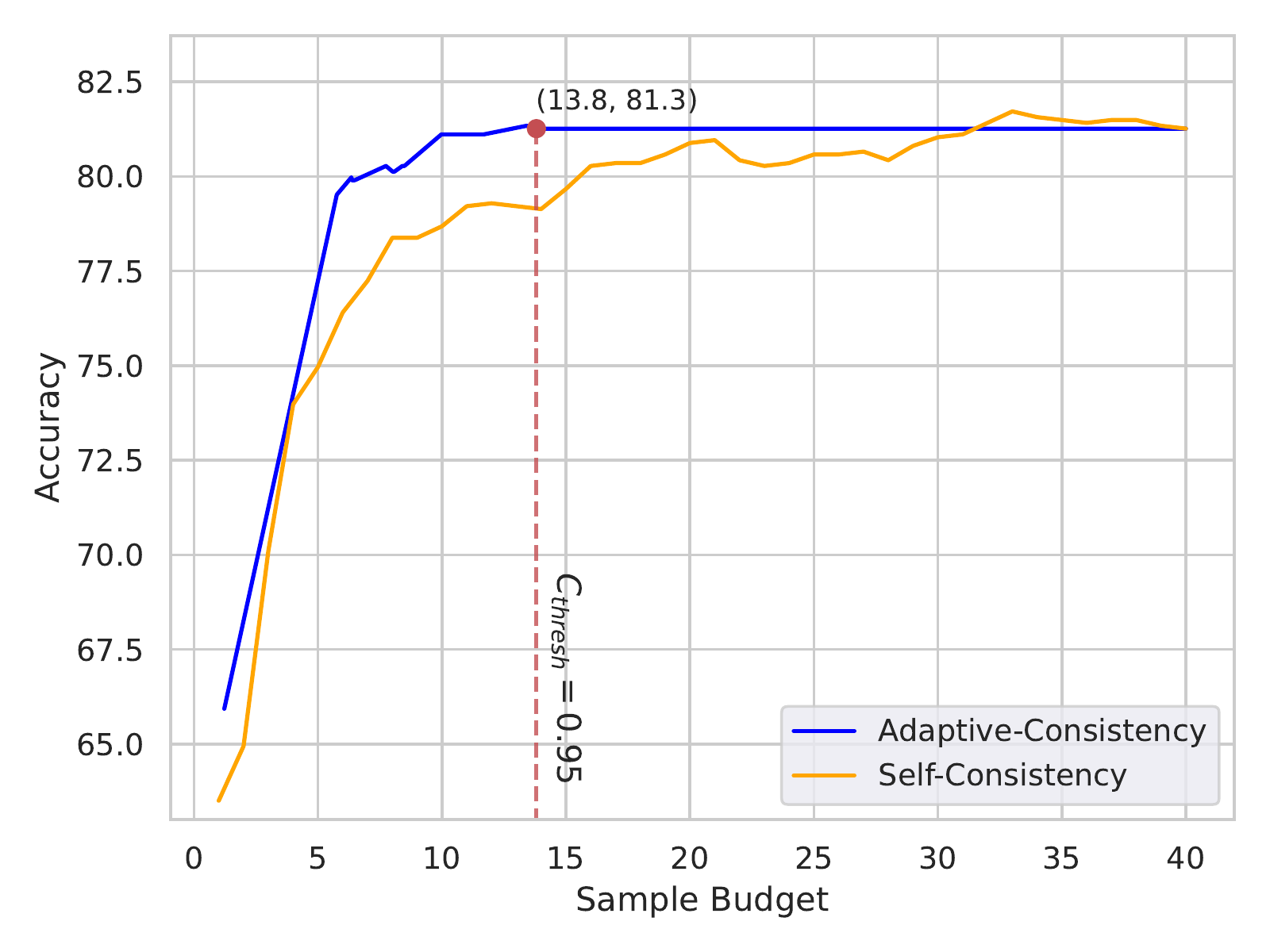}}
  \subfigure[\asdiv]{\includegraphics[width=0.32\linewidth]{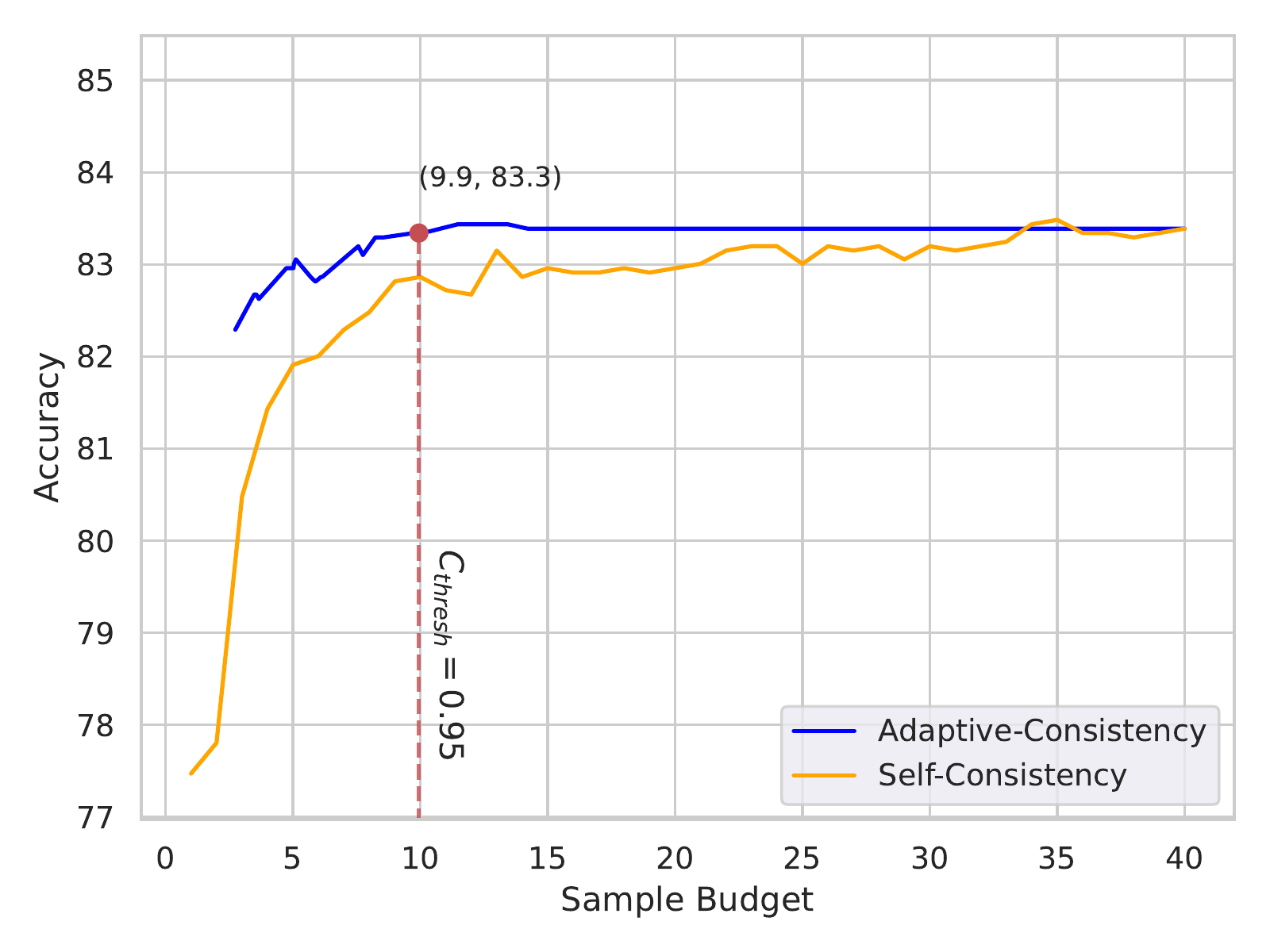}}
  \subfigure[\svamp]{\includegraphics[width=0.32\linewidth]{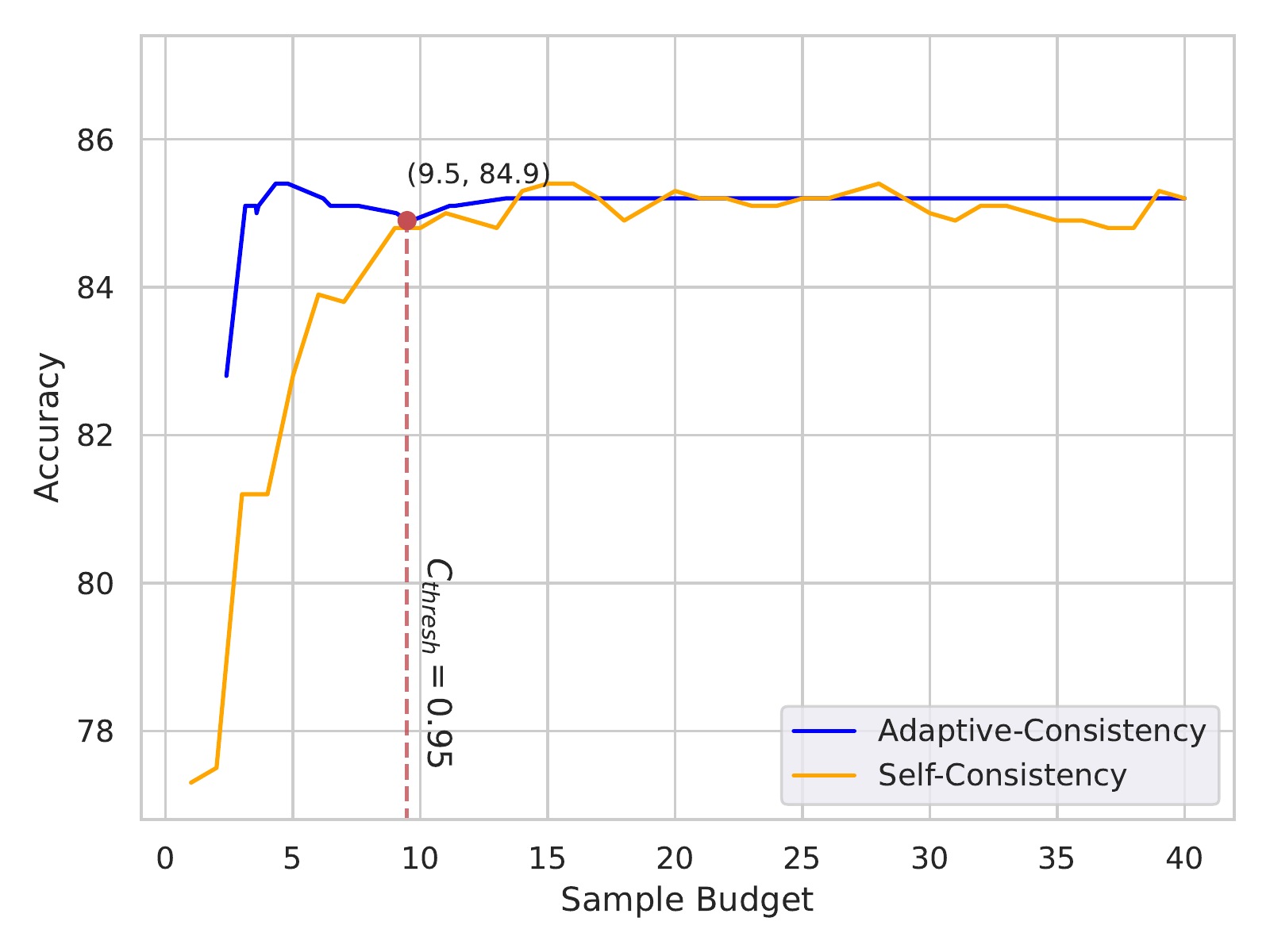}}
  \subfigure[\dateunderstanding]{\includegraphics[width=0.32\linewidth]{figures/date_cost_quality.pdf}}
  \subfigure[\trackingthree]{\includegraphics[width=0.32\linewidth]{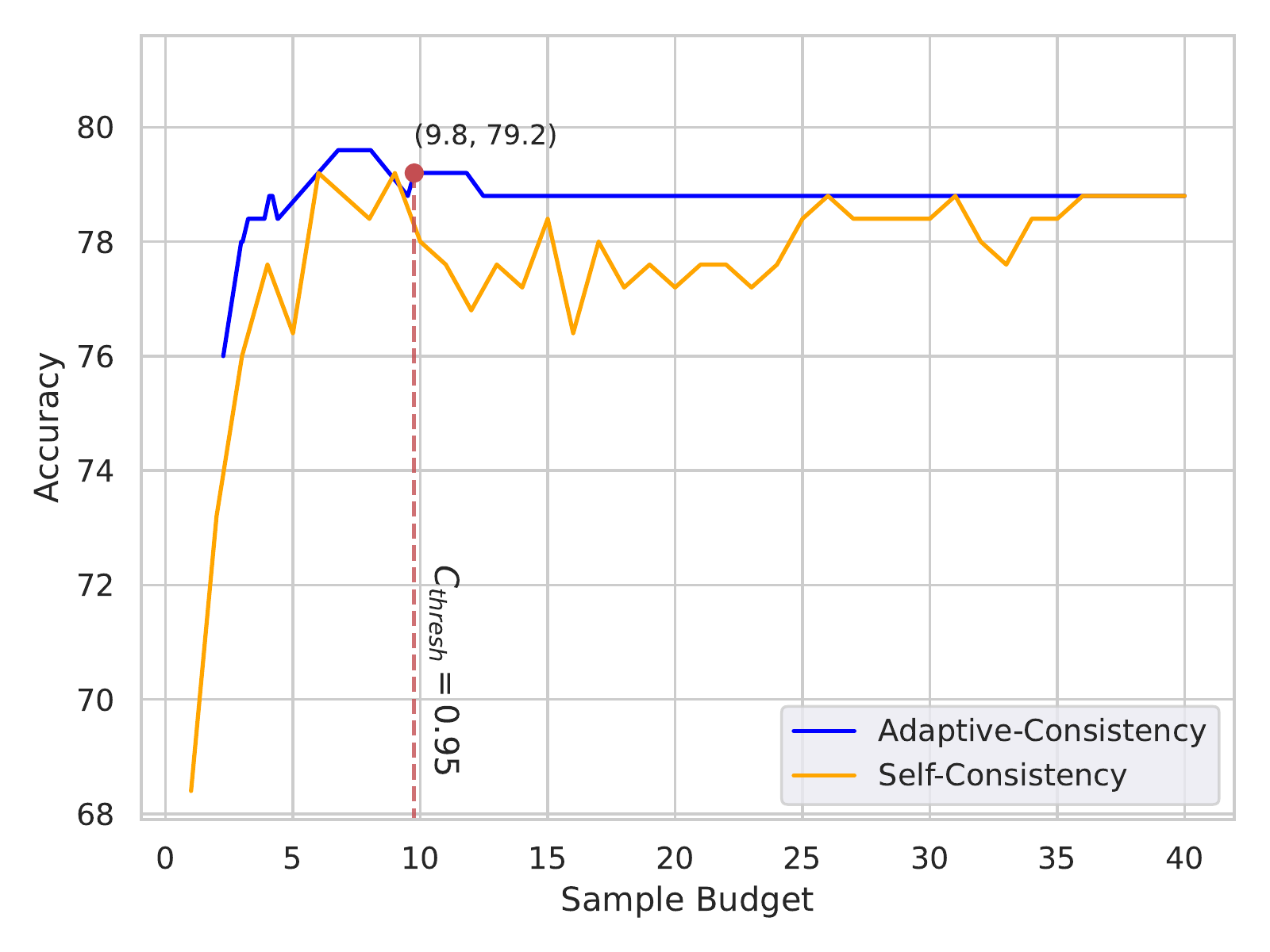}}
  \subfigure[\logicalthree]{\includegraphics[width=0.32\linewidth]{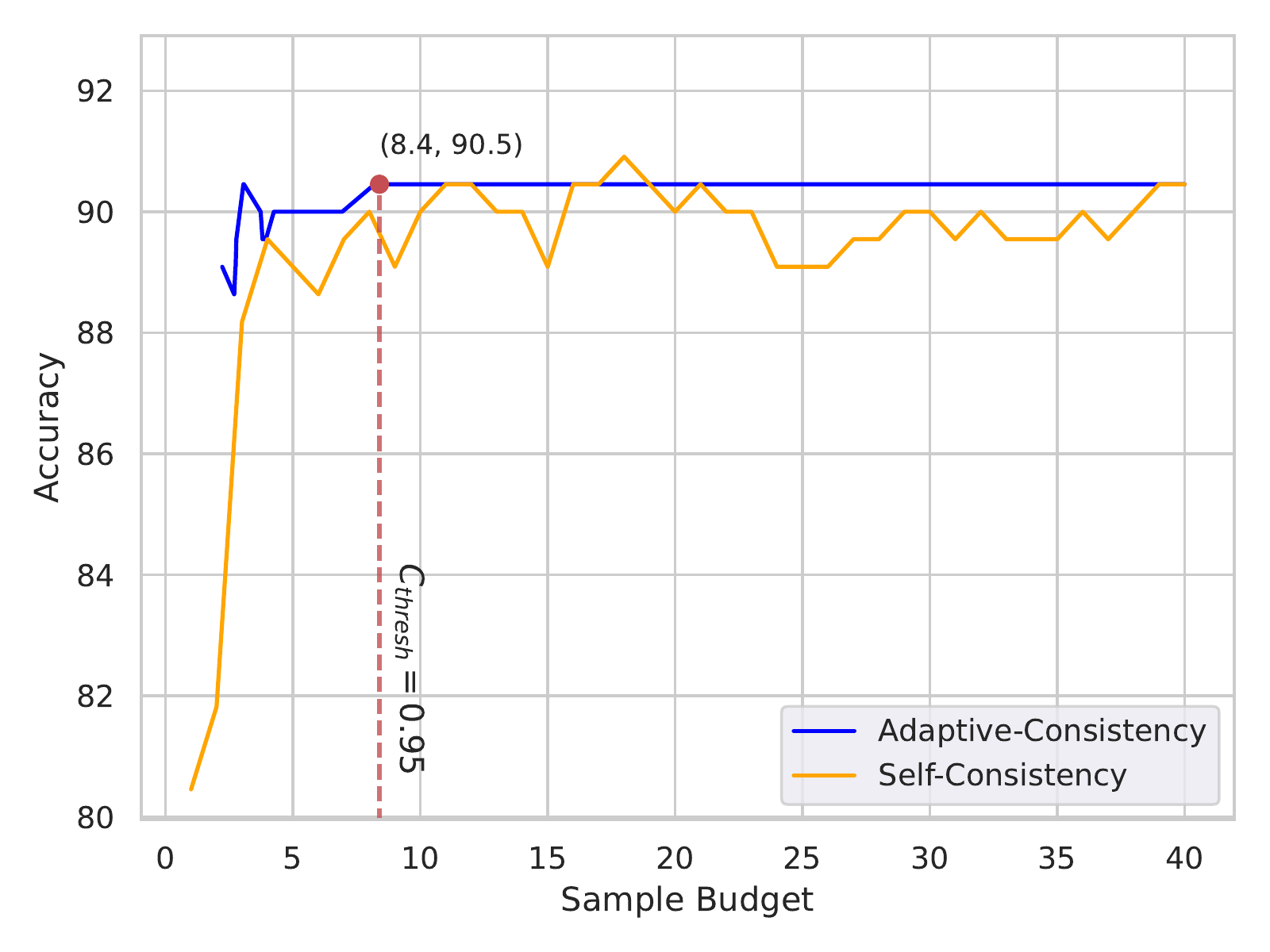}}
  \subfigure[\strategyqa]{\includegraphics[width=0.32\linewidth]{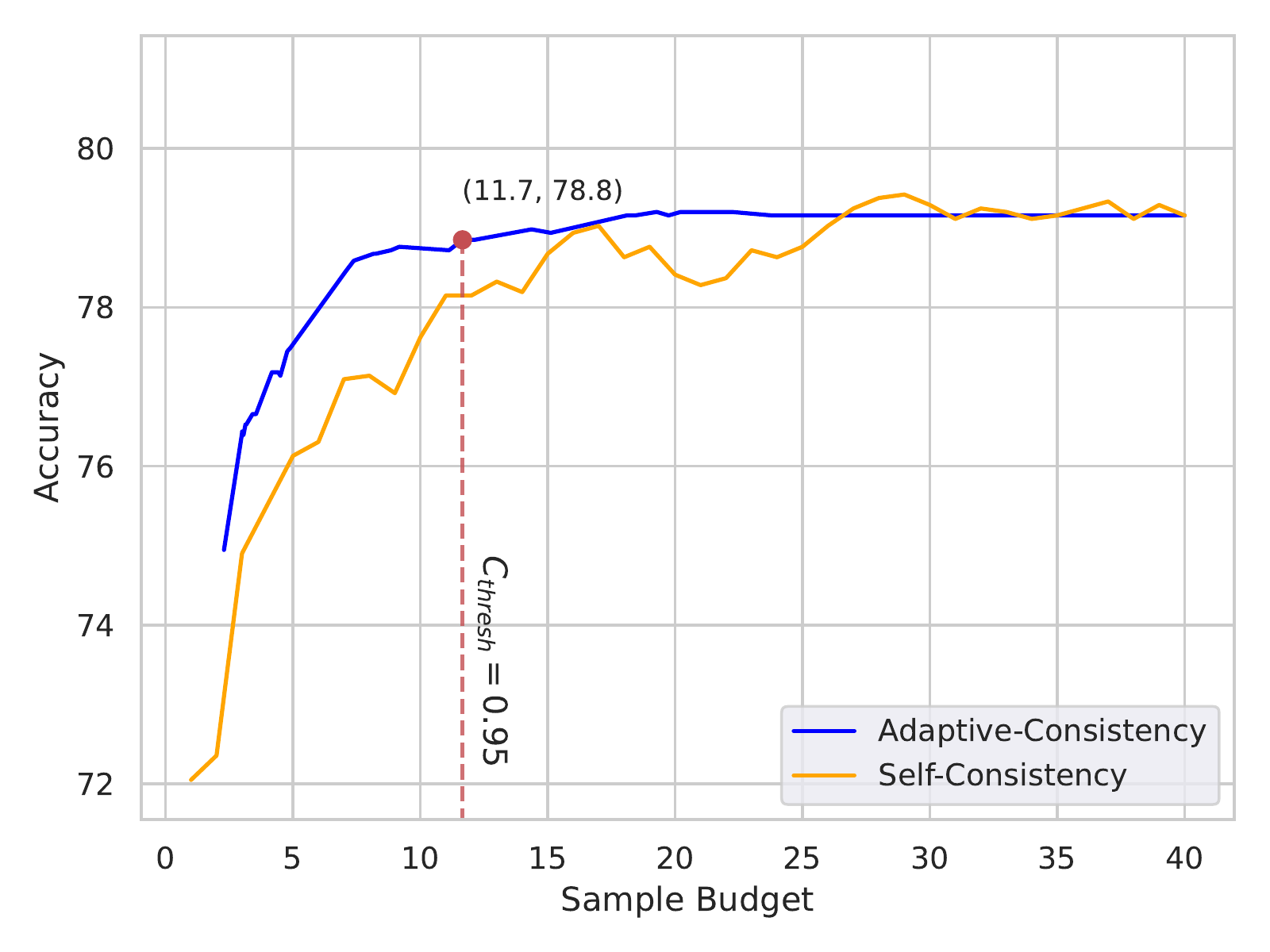}}
  \subfigure[\booleanexpressions]{\includegraphics[width=0.32\linewidth]{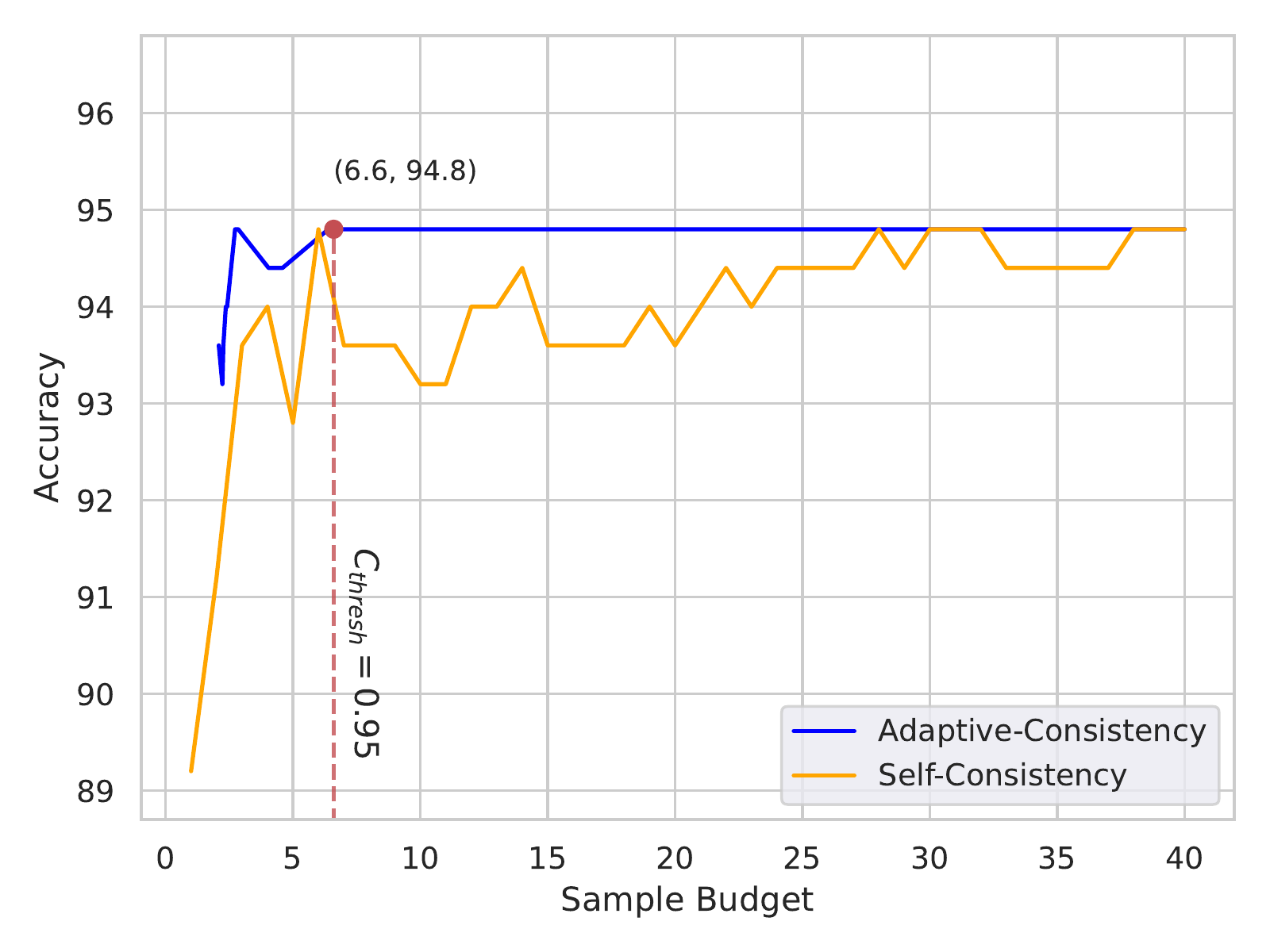}}
  \subfigure[\snarks]{\includegraphics[width=0.32\linewidth]{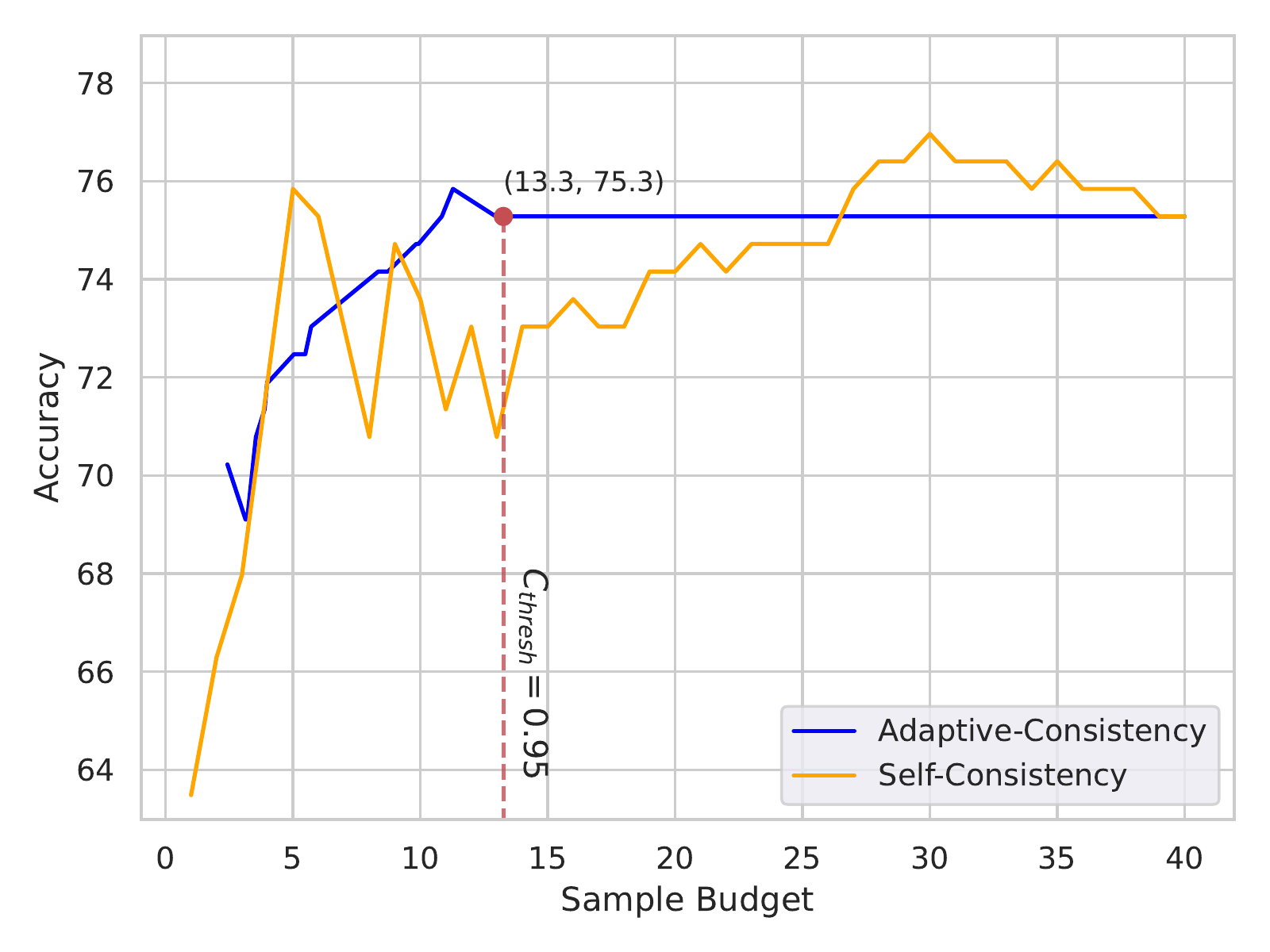}}
  \subfigure[\ruinnames]{\includegraphics[width=0.24\linewidth]{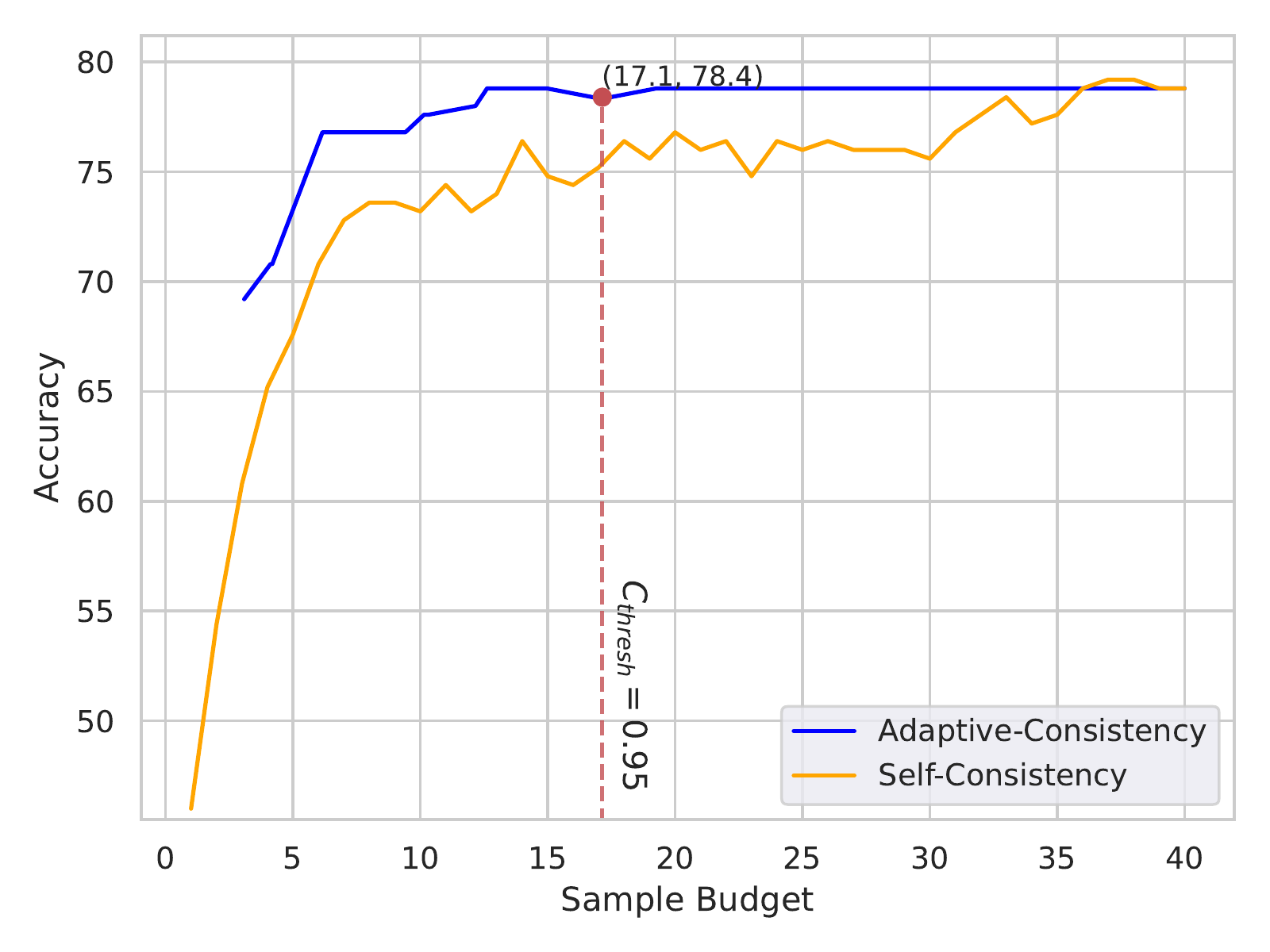}}
  \subfigure[\salienttranslation]{\includegraphics[width=0.24\linewidth]{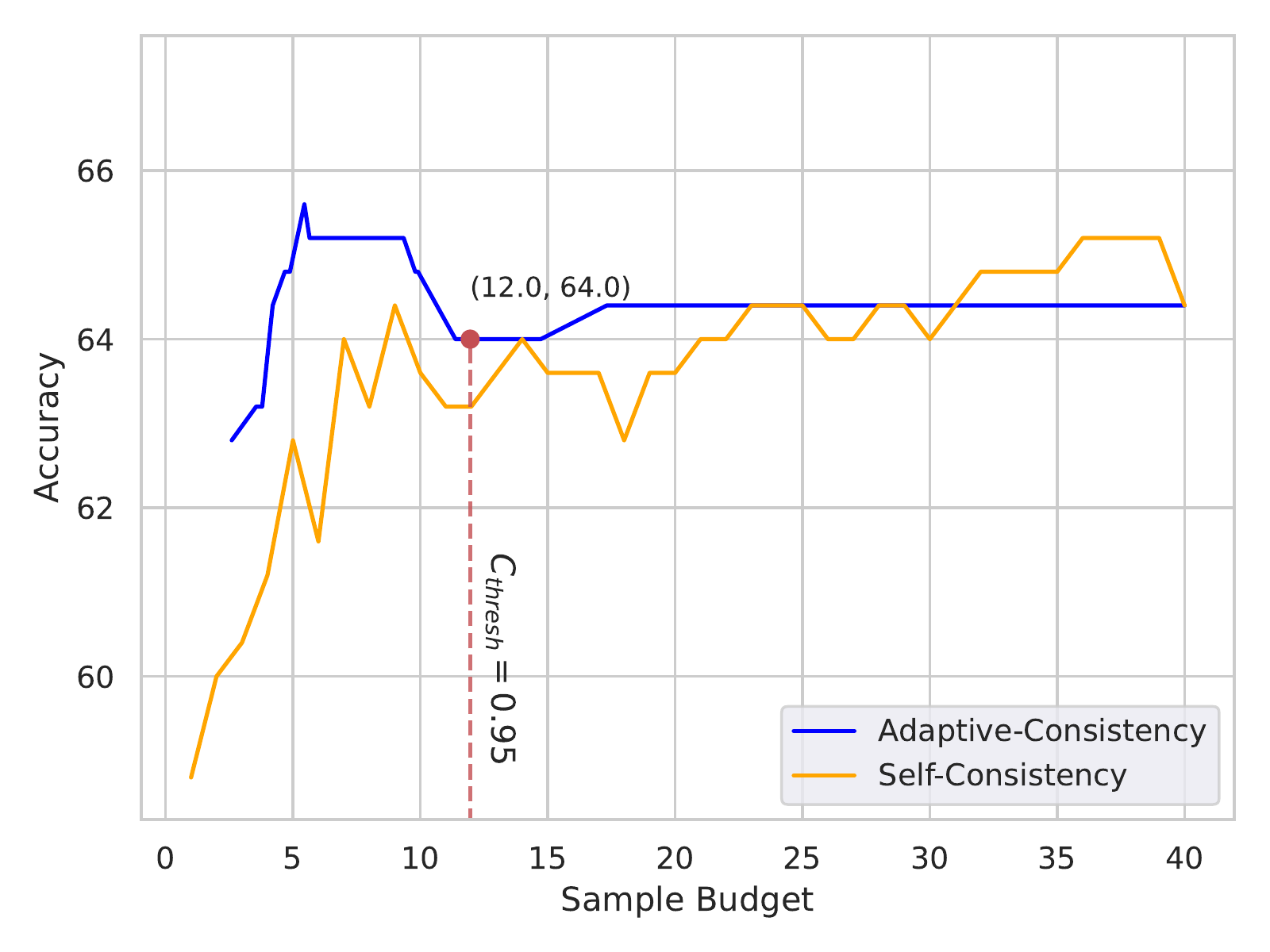}}
  \subfigure[\disambiguationqa]{\includegraphics[width=0.24\linewidth]{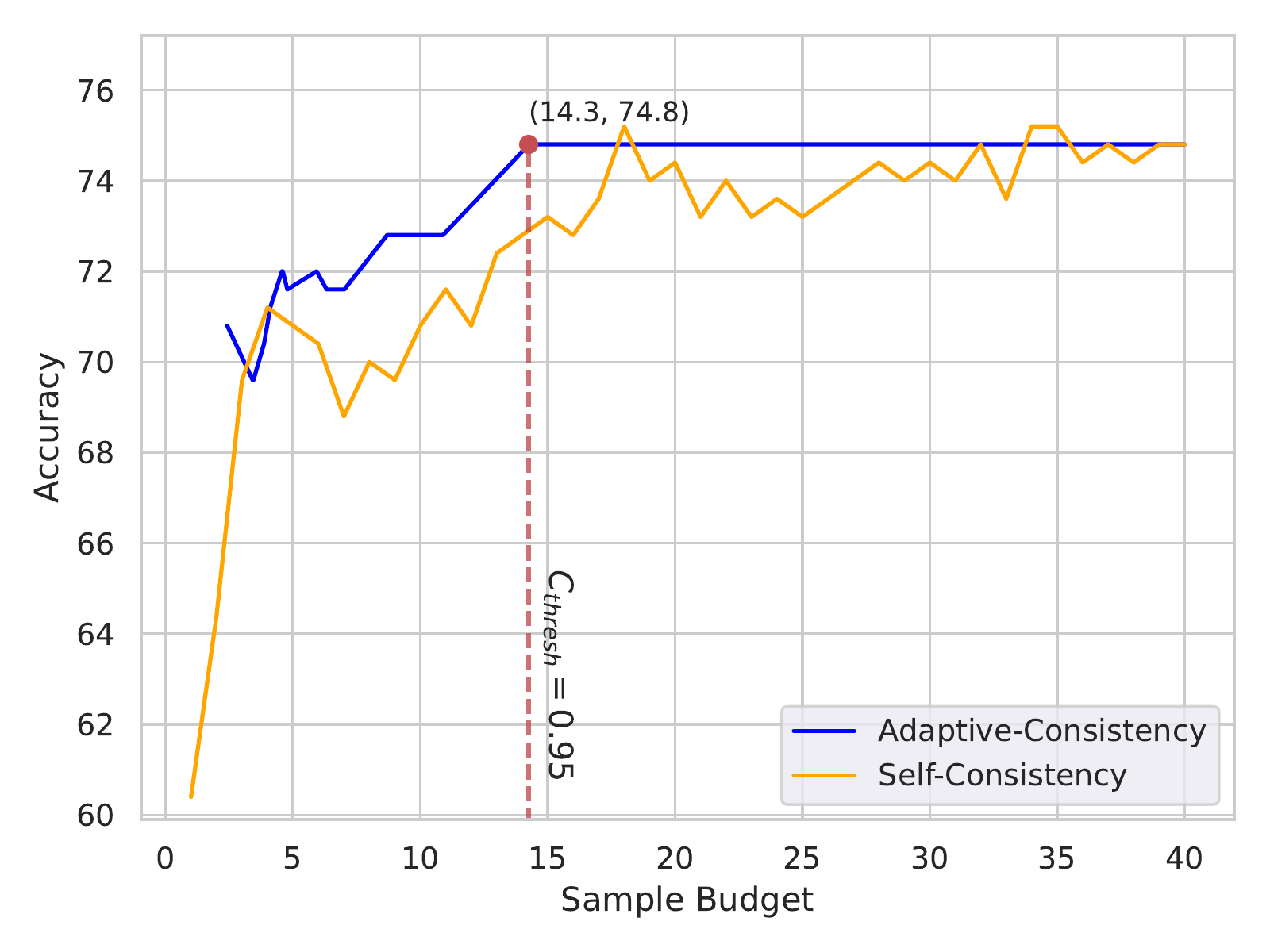}}
  \subfigure[\penguins]{\includegraphics[width=0.24\linewidth]{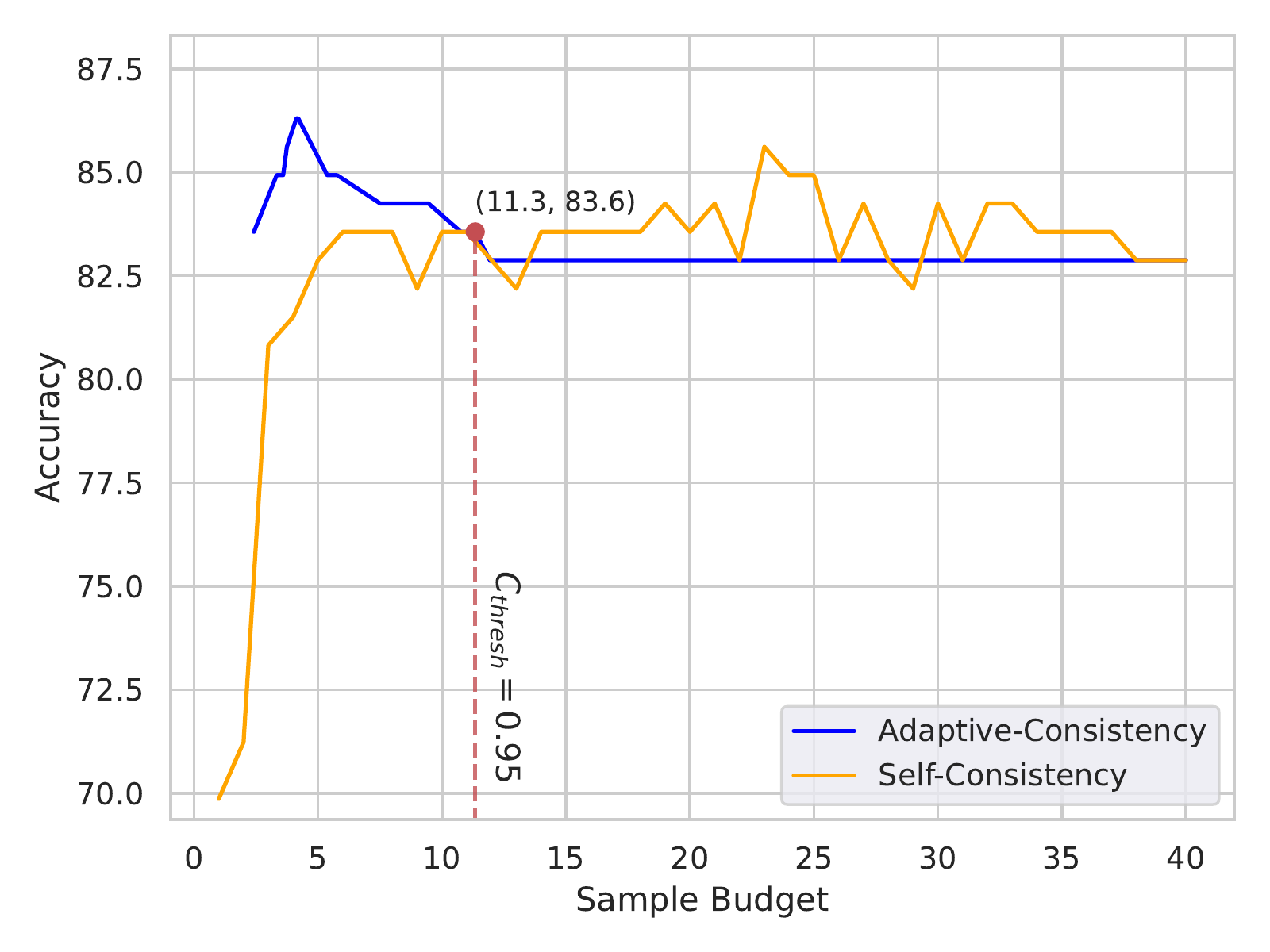}}

  \caption{Comparison of \ours with \self on various average sampling costs. \ours is able to consistently beat \self, especially when the sampling cost is low. Moreover, $C_{thresh} = 0.95$ is a good indication of saturation in accuracy indicating the value works out-of-box for most configurations considered.}
  \label{fig:cost_quality_complete}
\end{figure*}

In Section~\ref{anlys:cost_quality}, we demonstrate that \ours achieve better accuracy over \self when both are operating on same expected sample cost. In Figure~\ref{fig:cost_quality_complete} we show the complete results. 

Section~\ref{sec:results} previously demonstrated that \ours achieves comparable performance to \self using fewer samples. In this section, we consider a scenario where \ours and \self operate with the same average number of samples. For each fixed sampling budget $k$ of \self, we contrast the performance of \ours and \self, where \ours uses $k$ samples on average, rather than consistently across all instances.

Figure~\ref{fig:mathgraphs} provides a visual comparison of the performance of \ours and \self on \gsm: \ours outperforms \self in accuracy across all average sample costs. For example, when the average sample cost is 10, \ours achieves approximately 3\% higher accuracy on \gsm.

The success of \ours can be attributed to its adaptive sampling strategy. By varying the number of samples based on the complexity of the instance—using more samples where a clear consensus is hard to reach and fewer where answers are consistent—\ours manages to secure improved overall performance even when the average sample cost matches that of \self.

\subsection{\Stopfunc{}s}
\label{app:stopfuncs}

This section follows from the main discussion in Section~\ref{analy:stopfuncs}. We evaluate different \stopfunc{}s for \ours. We evaluate 6 different functions:
\begin{enumerate}
    \item \random{}: randomly stopping with a probability $p$, 2.) 
    \item \majority{}: stopping after the most common answer has a majority above a threshold, 
    \item \entropy{}: stopping after the entropy of answers is below a threshold, 
    \item \binomial: The main stopping criteria used in \ours, based on the Equation~\eqref{eq:betaeq}, 
    \item \dirichlet: The stopping criteria, based on Equation~\eqref{eq:dirichprob}.
    \item \Crp (\crp): The stopping criteria, which models probability as chinese restaurant process making no assumption on possible number of unique answers.
    
\end{enumerate}

\begin{figure*}[h]
  \centering
  \subfigure[\gsm]{\includegraphics[width=0.49\linewidth]{figures/gsm_beta_vs_entropy.pdf}}
  \subfigure[\strategyqa]{\includegraphics[width=0.49\linewidth]{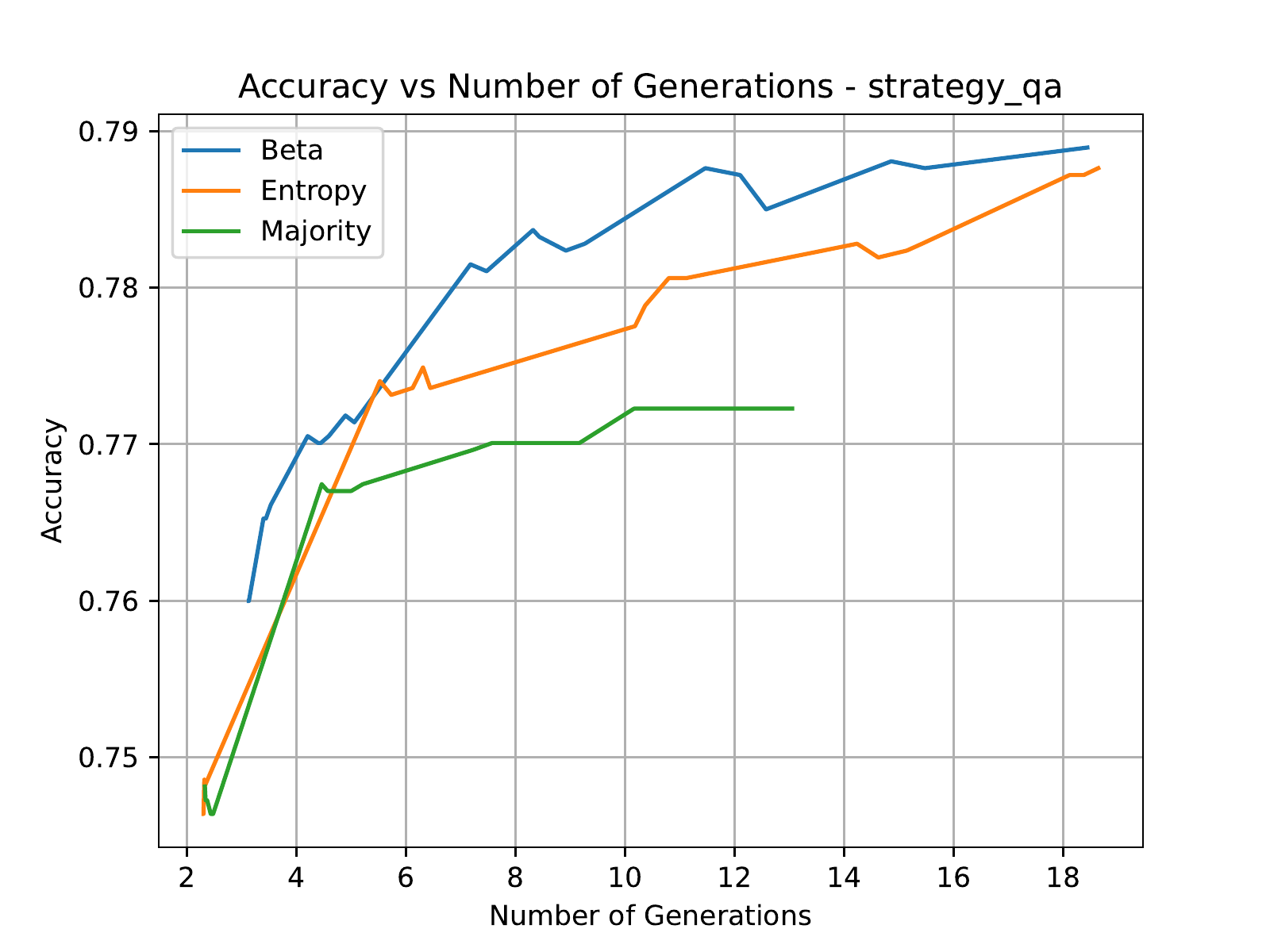}}
  \caption{Comparison of \binomial, \majority{} and \entropy \stopfunc{}s. In the two representative datasets, \binomial consistently beats \entropy and \majority for the same sampling cost. This shows in practice \binomial performs better than both for the desirable range of accuracy and sampling cost.}
  \label{app:betavsentropy}
\end{figure*}

For comparison, we tune the $C_{thresh}$ in each case on the training set of \gsm dataset. Results are presented in Table~\ref{tab:stoptable}. \random and \majority{} are inferior to \binomial across all datasets and models. Further, while \dirichlet and \crp are almost similar to \binomial, they are relatively very slow. While  Although, from Table~\ref{tab:stoptable}, \entropy{} looks appears to be on par with \binomial, in Figure~\ref{app:betavsentropy}, we show \binomial beats \entropy given the same expected sampling cost. 

Finally, \binomial has additional key advantages: \binomial incorporates a measure of uncertainty, which makes it more robust to variations in data order, mitigates the influence of noise, and offers a quantitative measure of confidence in the majority outcome. 
Consider an extreme case where the first two generated solutions are identical. The majority voting strategy would instantly halt the process, potentially missing out on better solutions. In contrast, \binomial will keep sampling as the confidence for stopping has not yet reached.

\subsection{Chinese Restaurant Process}
\label{app:crp}

In the \dirichlet \stopfunc{}, we assume that the number of unique answers that can be generated by the \llm is known in advance (and equal to the number of unique answers in the current observation set). However, this assumption may not hold for datasets such as \gsm, where numerical answers are expected. The \Crp (\crp) is a generalization of the \dirichlet process that addresses this limitation by not making any assumption on the number of unique answers. 

In \crp, we consider a list of same answers as a cluster, denoted by $c_i$, where $i$ is the index of the cluster. Let $n_i$ be the number of elements in cluster $c_i$, and $n$ be the total number of elements across all clusters. The probability of a new answer belonging to an existing cluster $c_i$ is directly proportional to the size of the cluster, and is given by:
\begin{equation}
P(c_i) = \frac{n_i}{n + \alpha},
\end{equation}
whereas the probability that a new unseen answer will form a new cluster is given by:
\begin{equation}
P(c_{new}) = \frac{\alpha}{n + \alpha},
\end{equation}
where $\alpha$ is the concentration parameter, which parameterizes the probability of generating a new answer.

Our goal is to calculate the probability that the current majority cluster in observations will remain the same even with more generations. The first task is to estimate the concentration parameter $\alpha$. We use the approximation proposed by \cite{West1992HyperparameterEI} to model the $\alpha$ as
\begin{equation}
p(\alpha|k,n) \approx G(a+k-1, b+\gamma+\log(n)),
\end{equation}
where $k$ is the number of unique answers (clusters) in the current observation, $n$ is the total number of answers, $a$ and $b$ are priors and both set equal to 1, and $\gamma$ is Euler's constant and $G(\alpha; a+k-1, b+\gamma+\log(n))$ denotes the probability density function of the Gamma distribution with shape parameter $a+k-1$ and rate parameter $b+\gamma+\log(n)$.

We sample $\alpha$ multiple times (100), and for each sample, we run Monte-Carlo Simulation (1000 simulations) based on the CRP probability modeling. Each simulation starts from from current set of observations, and performed till 40 generations are sampled. The probability that the current majority cluster remains the majority is then given by:

\begin{equation}
P(\text{majority}) = \frac{1}{N_\alpha N_{MCS}} \sum_{i=1}^{N_\alpha} \sum_{j=1}^{N_{MCS}} I(\text{majority}_n^{40}),
\end{equation}
where $N_\alpha$ is the number of times we sample $\alpha$, $N_{MCS}$ is the number of Monte-Carlo Simulations, and $I(\text{majority}_n^{40})$ is an indicator function that equals 1 if the current majority remains the majority after $40$ generations, and 0 otherwise.

\begin{table*}[h]
  \centering
  \resizebox{1\linewidth}{!}{
    \begin{tabular}{@{}lc|cc|cc|cc|cc|cc|cc@{}}
    \toprule
     & & \multicolumn{2}{c|} {\textbf{\random}} & \multicolumn{2}{c|}{\textbf{\majority}} & \multicolumn{2}{c}{\entropy} & \multicolumn{2}{c}{\binomial (\ours)} &   \multicolumn{2}{c}{\dirichlet} & \multicolumn{2}{c}{\crp} \\
     \cmidrule(lr){3-4} \cmidrule(lr){5-6} \cmidrule(lr){7-8} \cmidrule(lr){9-10} \cmidrule(lr){11-12} \cmidrule(lr){13-14}
     & & \textbf{Average $\downarrow$} & \multirow{2}{*}{\textbf{{Accuracy $\uparrow$}}} & \textbf{Average $\downarrow$} & \multirow{2}{*}{\textbf{{Accuracy $\uparrow$}}} & \textbf{Average $\uparrow$} & \multirow{2}{*}{\textbf{{Accuracy $\uparrow$}}} & \textbf{Average $\uparrow$} & \multirow{2}{*}{\textbf{{Accuracy $\uparrow$}}} & \textbf{Average $\uparrow$} & \multirow{2}{*}{\textbf{{Accuracy $\uparrow$}}} & \textbf{Average $\uparrow$} & \multirow{2}{*}{\textbf{{Accuracy $\uparrow$}}} \\
     & & \textbf{{Generations }} & & \textbf{{Generations }} & & \textbf{{Generations }} & & \textbf{{Generations }} & & \textbf{{Generations }} & & \textbf{{Generations }} &\\
    \midrule

\addlinespace

\multirow{2}{*}{\textbf{\gsm}} & \vicuna & $\textbf{26.0}$ & $30.1$ & $28.7$ & $31.5$ & $26.3$ & $31.5$ & $26.8$ & $31.5$ & $28.2$ & $\textbf{31.7}$& $\textbf{25.6}$ & $31.5$ \\

\addlinespace

& \codex & $13.8$ & $76.9$ & $16.6$ & $80.9$ & $15.3$ & $81.0$ & $13.8$ & $81.0$ & $15.2$ & $\textbf{81.1}$& $\textbf{13.2}$ & $\textbf{81.1}$ \\

\addlinespace

\hline

\multirow{2}{*}{\textbf{\asdiv}} & \vicuna & $28.0$ & $63.2$ & $\textbf{14.8}$ & $63.7$ & $15.8$ & $63.9$ & $16.5$ & $\textbf{64.0}$ & $17.7$ & $\textbf{64.0}$& $16.9$ & $\textbf{64.0}$ \\

\addlinespace

& \codex & $13.8$ & $81.9$ & $\textbf{9.2}$ & $83.1$ & $11.5$ & $\textbf{83.3}$ & $10.0$ & $83.2$ & $10.7$ & $83.1$& $10.7$ & $83.1$ \\

\addlinespace

\hline

\multirow{2}{*}{\textbf{\svamp}} & \vicuna & $28.0$ & $61.3$ & $\textbf{17.1}$ & $62.5$ & $17.7$ & $62.6$ & $18.8$ & $62.8$ & $19.7$ & $\textbf{62.9}$& $18.2$ & $62.8$ \\

\addlinespace

& \codex & $13.4$ & $83.3$ & $\textbf{8.4}$ & $84.8$ & $10.7$ & $85.1$ & $9.5$ & $85.0$ & $10.3$ & $\textbf{85.1}$& $9.8$ & $85.0$ \\

\addlinespace

\hline

\multirow{2}{*}{\textbf{\dateunderstanding}} & \vicuna & $28.0$ & $58.3$ & $\textbf{15.3}$ & $59.5$ & $16.0$ & $59.9$ & $17.3$ & $\textbf{60.2}$ & $18.5$ & $59.9$& $16.9$ & $59.9$ \\

\addlinespace

& \codex & $13.2$ & $76.4$ & $\textbf{9.7}$ & $78.7$ & $11.6$ & $79.9$ & $10.7$ & $79.5$ & $11.9$ & $\textbf{80.5}$& $10.7$ & $79.8$ \\

\addlinespace

\hline

\multirow{2}{*}{\textbf{\trackingthree}} & \vicuna & $27.9$ & $31.8$ & $\textbf{15.0}$ & $\textbf{33.0}$ & $18.4$ & $32.0$ & $20.3$ & $32.0$ & $23.3$ & $32.0$& $19.6$ & $31.8$ \\

\addlinespace

& \codex & $13.5$ & $76.3$ & $\textbf{7.0}$ & $76.8$ & $11.5$ & $76.9$ & $9.7$ & $77.1$ & $11.5$ & $\textbf{77.2}$ & $10.2$ & $77.1$ \\

\addlinespace

\addlinespace

\hline

\multirow{2}{*}{\textbf{\logicalthree}} & \vicuna & $27.9$ & $50.5$ & $\textbf{12.9}$ & $51.2$ & $15.8$ & $\textbf{51.4}$ & $18.1$ & $\textbf{51.4}$ & $20.9$ & $51.2$& $18.3$ & $\textbf{51.4}$ \\

\addlinespace

& \codex & $13.7$ & $88.3$ & $\textbf{5.9}$ & $\textbf{89.6}$ & $10.1$ & $\textbf{89.6}$ & $8.5$ & $89.4$ & $10.2$ & $89.2$& $9.3$ & $89.4$ \\

\addlinespace

\hline

\multirow{2}{*}{\textbf{\strategyqa}} & \vicuna & $28.1$ & $65.1$ & $\textbf{11.7}$ & $65.5$ & $14.5$ & $65.8$ & $16.3$ & $65.8$ & $18.7$ & $\textbf{65.8}$& $17.0$ & $65.7$ \\

\addlinespace

& \codex & $13.4$ & $76.6$ & $\textbf{7.2}$ & $77.8$ & $14.9$ & $78.5$ & $11.9$ & $78.8$ & $14.5$ & $\textbf{78.9}$& $11.4$ & $\textbf{78.9}$ \\

\addlinespace

\hline

\multirow{2}{*}{\textbf{\booleanexpressions}} & \vicuna & $27.6$ & $78.0$ & $\textbf{10.4}$ & $76.8$ & $14.8$ & $78.3$ & $16.2$ & $78.4$ & $19.1$ & $\textbf{78.8}$& $17.0$ & $78.5$ \\

\addlinespace

& \codex & $13.1$ & $93.4$ & $\textbf{4.3}$ & $94.3$ & $8.2$ & $\textbf{94.5}$ & $6.6$ & $\textbf{94.5}$ & $8.2$ & $\textbf{94.5}$& $7.9$ & $94.4$ \\

\addlinespace

\hline

\multirow{2}{*}{\textbf{\snarks}} & \vicuna & $28.4$ & $70.3$ & $\textbf{18.1}$ & $72.1$ & $20.3$ & $73.0$ & $23.2$ & $\textbf{73.6}$ & $25.8$ & $\textbf{73.6}$& $22.9$ & $\textbf{73.6}$ \\

\addlinespace

& \codex & $13.6$ & $71.6$ & $\textbf{10.5}$ & $\textbf{74.0}$ & $12.1$ & $\textbf{74.0}$ & $12.7$ & $\textbf{74.0}$ & $14.2$ & $73.4$& $12.3$ & $73.2$ \\

\addlinespace

\hline

\multirow{2}{*}{\textbf{\ruinnames}} & \vicuna & $\textbf{28.3}$ & $40.6$ & $30.4$ & $43.9$ & $31.9$ & $43.7$ & $33.8$ & $43.6$ & $34.0$ & $43.6$& $32.0$ & $\textbf{44.0}$ \\

\addlinespace

& \codex & $\textbf{13.8}$ & $71.7$ & $17.5$ & $77.7$ & $18.6$ & $\textbf{78.1}$ & $17.2$ & $78.0$ & $17.6$ & $76.8$& $16.4$ & $\textbf{78.1}$ \\

\addlinespace

\hline

\multirow{2}{*}{\textbf{\salienttranslation}} & \vicuna & $24.9$ & $27.7$ & $\textbf{24.6}$ & $28.5$ & $26.3$ & $28.0$ & $28.7$ & $28.7$ & $29.4$ & $28.8$& $26.9$ & $\textbf{28.9}$ \\

\addlinespace

& \codex & $14.0$ & $62.5$ & $\textbf{9.9}$ & $\textbf{64.7}$ & $13.1$ & $64.3$ & $11.8$ & $64.3$ & $13.7$ & $64.1$& $11.7$ & $64.3$ \\

\addlinespace

\hline

\multirow{2}{*}{\textbf{\disambiguationqa}} & \vicuna & $27.9$ & $62.9$ & $\textbf{18.3}$ & $63.5$ & $20.1$ & $63.1$ & $22.8$ & $63.5$ & $25.4$ & $\textbf{63.9}$& $22.1$ & $63.3$ \\

\addlinespace

& \codex & $13.7$ & $72.1$ & $\textbf{10.4}$ & $73.9$ & $15.9$ & $74.9$ & $13.5$ & $75.1$ & $16.3$ & $\textbf{75.2}$& $13.2$ & $\textbf{75.2}$ \\

\addlinespace

\hline

\multirow{2}{*}{\textbf{\penguins}} & \vicuna & $27.9$ & $45.6$ & $\textbf{19.7}$ & $46.3$ & $20.7$ & $\textbf{47.3}$ & $22.9$ & $\textbf{47.3}$ & $25.1$ & $\textbf{47.3}$& $22.1$ & $\textbf{47.3}$ \\

\addlinespace

& \codex & $13.3$ & $81.4$ & $\textbf{9.0}$ & $83.3$ & $13.1$ & $83.8$ & $11.0$ & $84.0$ & $12.9$ & $84.0$& $11.0$ & $\textbf{84.5}$ \\

\addlinespace

\bottomrule
  \end{tabular}
  }
\caption{Comparison of various \Stopfunc{}s in \ours. In general, \binomial outperforms \random and \majority by decent margins across all datasets. \binomial has comparable performance to \dirichlet, but the latter is much slower. \entropy performs similarly to \binomial but lacks human-interpretable stopping rationale. }

 \label{tab:stoptable}
\end{table*}

\subsection{Effect of Confidence Threshold on \ours}
\label{app:confthresheffect}

We follow the discussion in Section~\ref{anlys:sampbudgets}, and present complete results on all datasets for \codex. 

\begin{figure*}[h]
  \centering
  \subfigure[\gsm]{\includegraphics[width=0.32\linewidth]{figures/gsm_sample_compare.pdf}}
  \subfigure[\asdiv]{\includegraphics[width=0.32\linewidth]{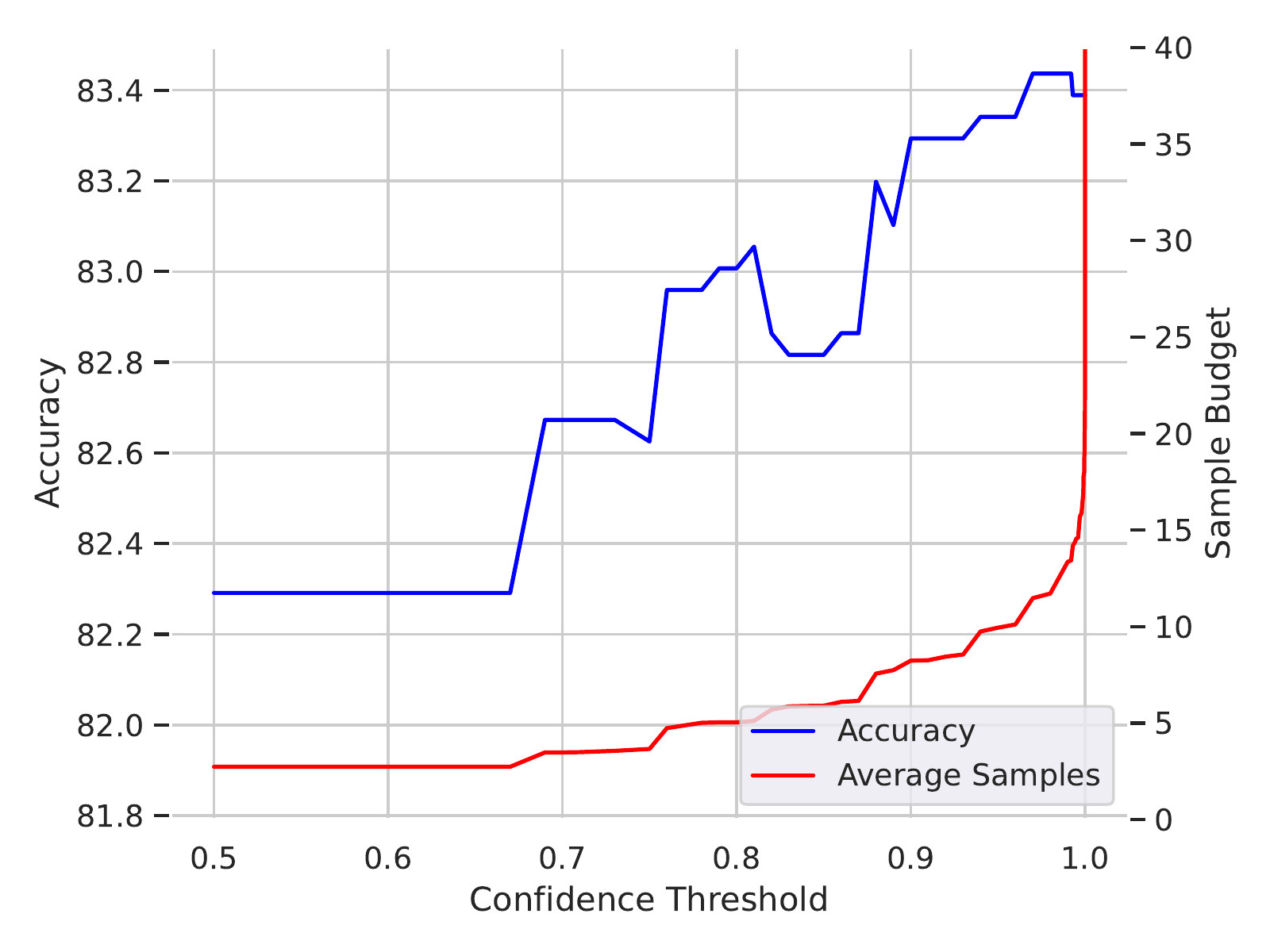}}
  \subfigure[\svamp]{\includegraphics[width=0.32\linewidth]{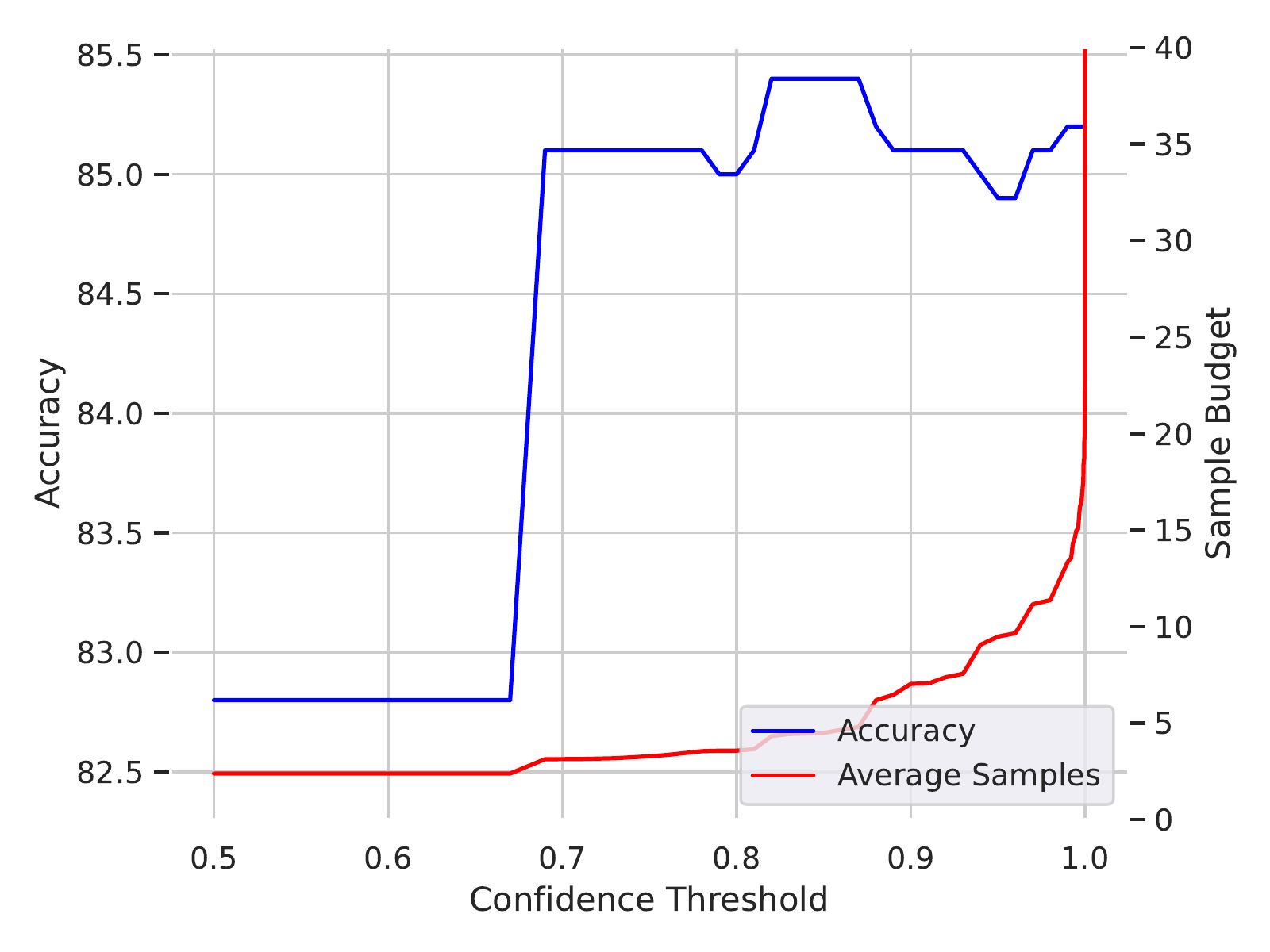}}
  \subfigure[\dateunderstanding]{\includegraphics[width=0.32\linewidth]{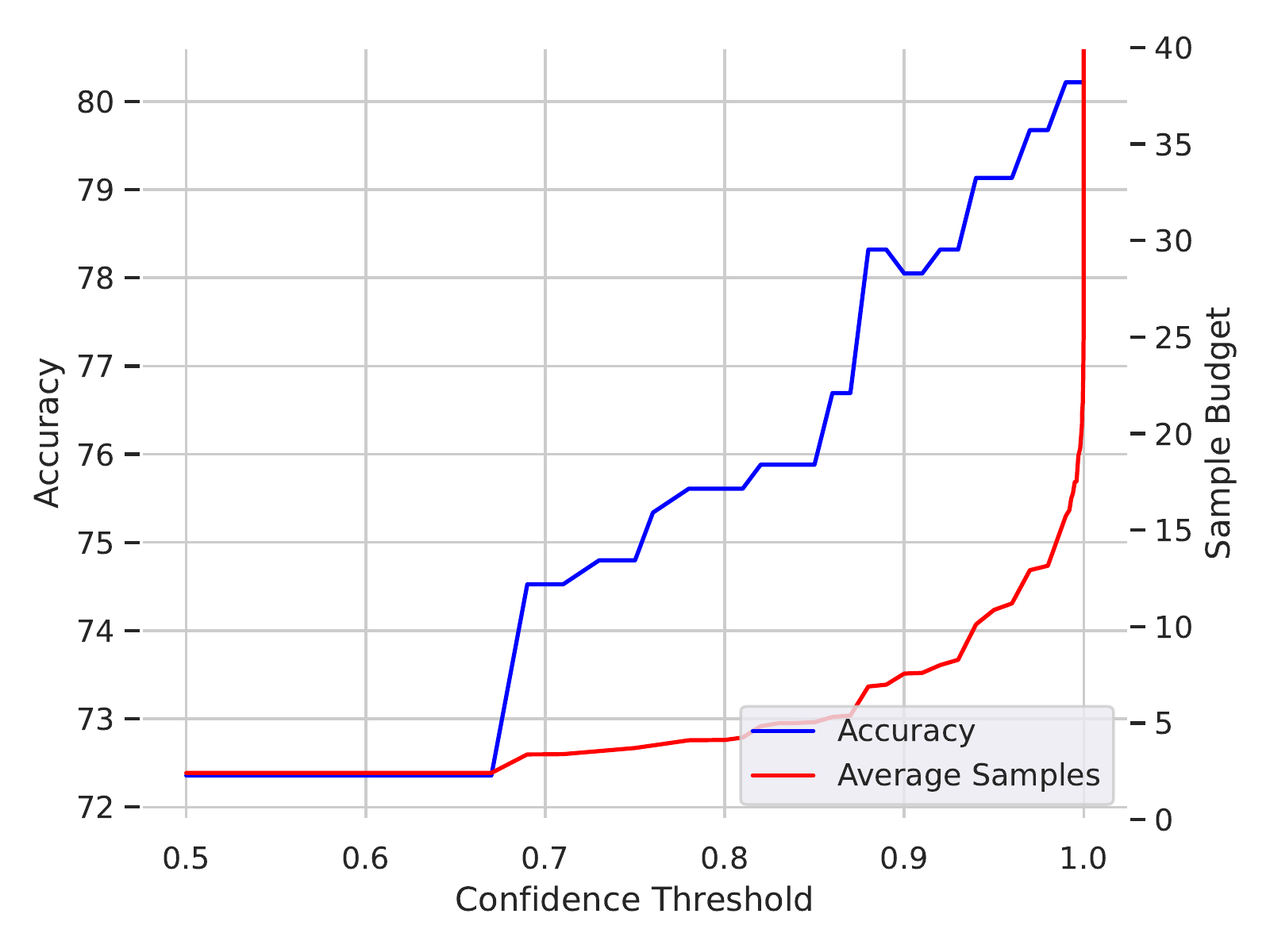}}
  \subfigure[\trackingthree]{\includegraphics[width=0.32\linewidth]{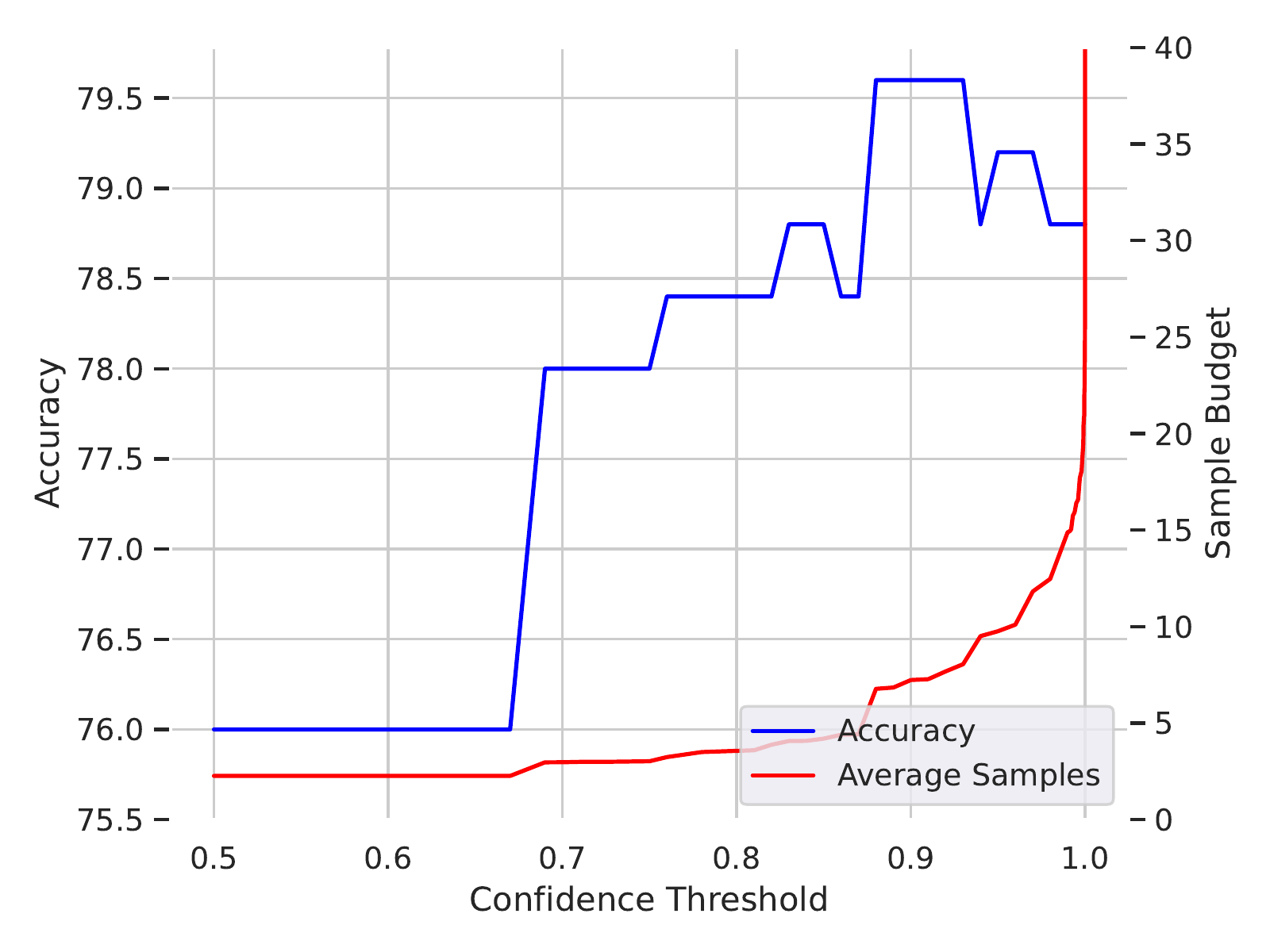}}
  \subfigure[\logicalthree]{\includegraphics[width=0.32\linewidth]{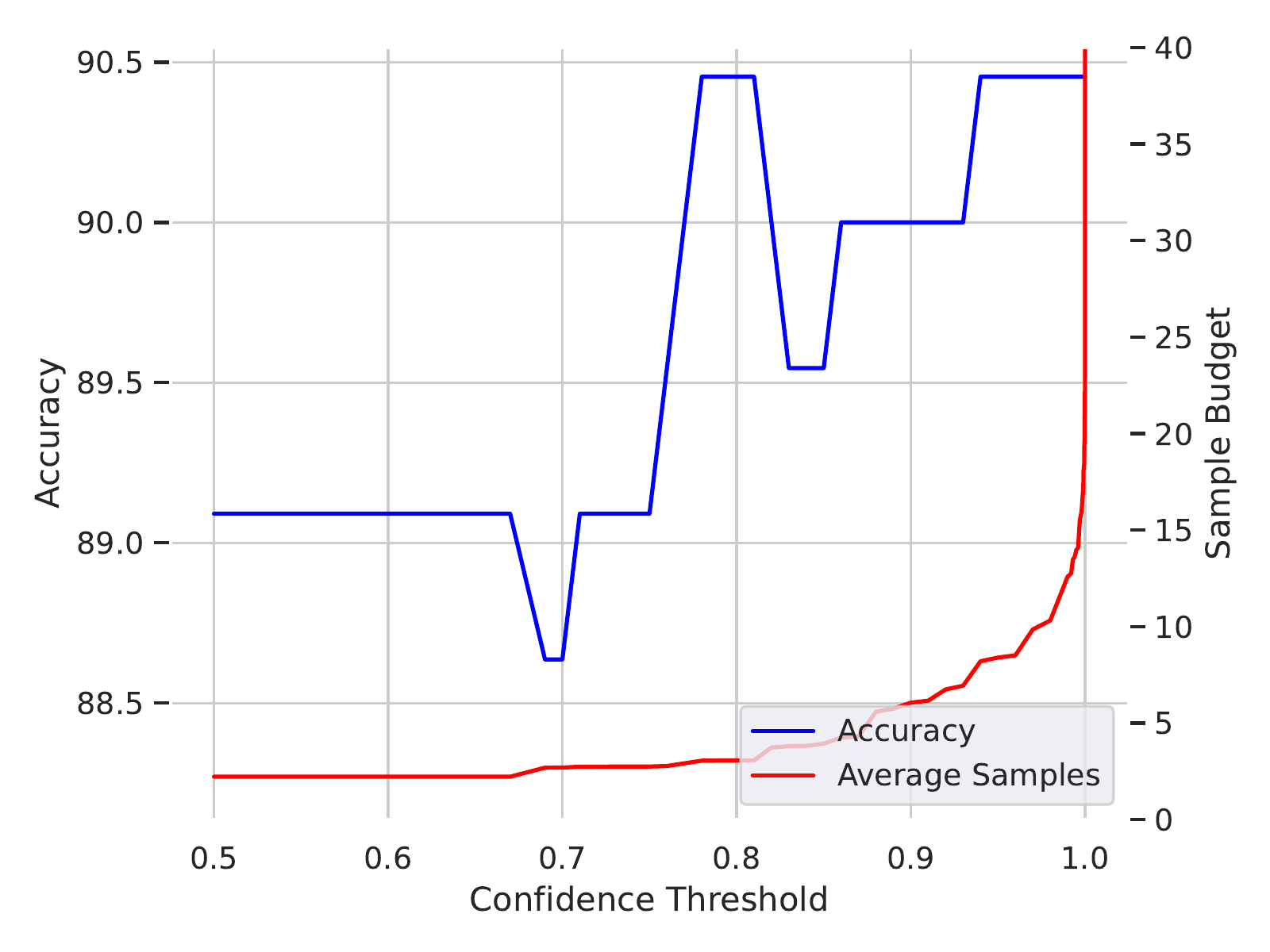}}
  \subfigure[\strategyqa]{\includegraphics[width=0.32\linewidth]{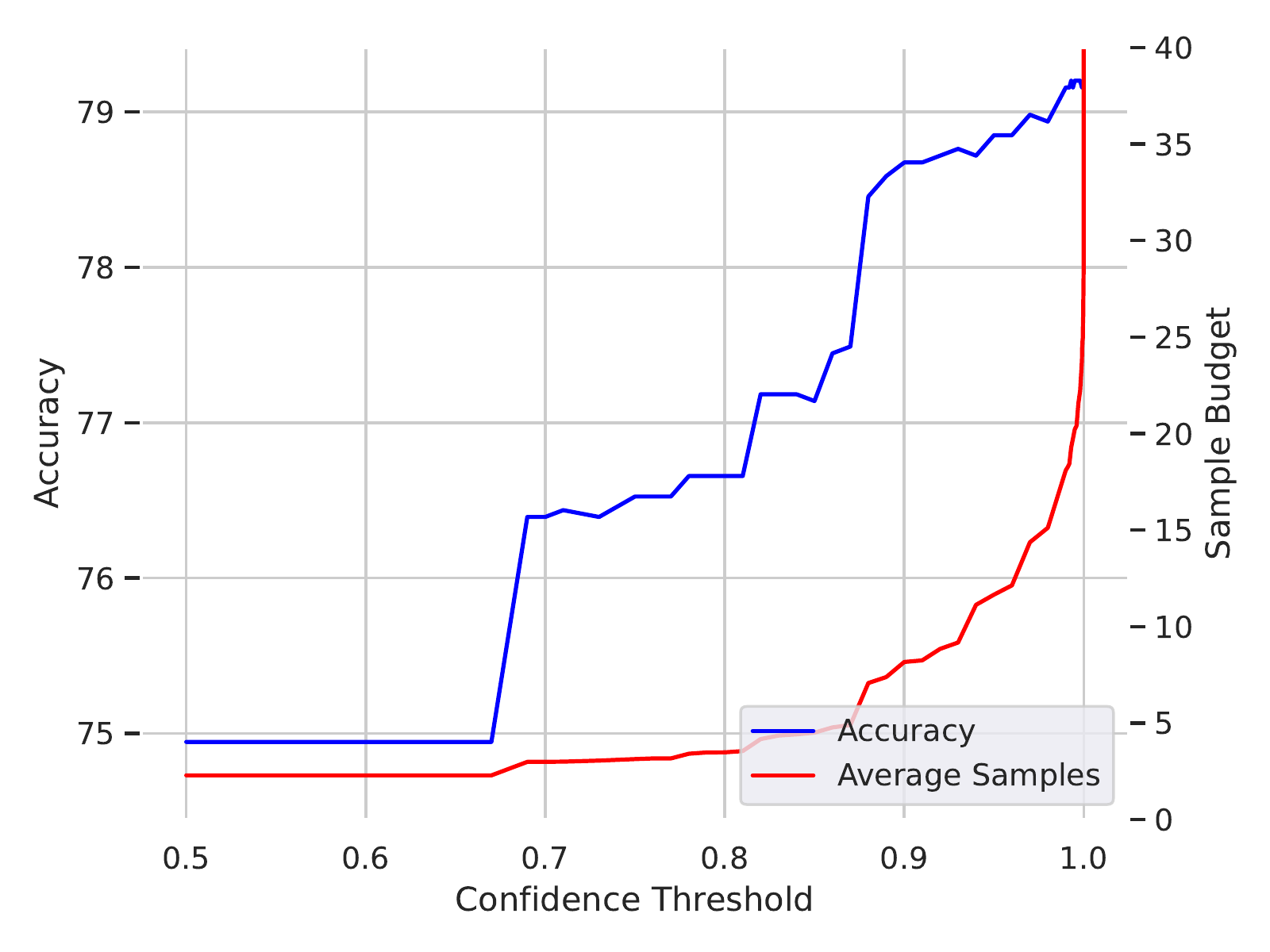}}
  \subfigure[\booleanexpressions]{\includegraphics[width=0.32\linewidth]{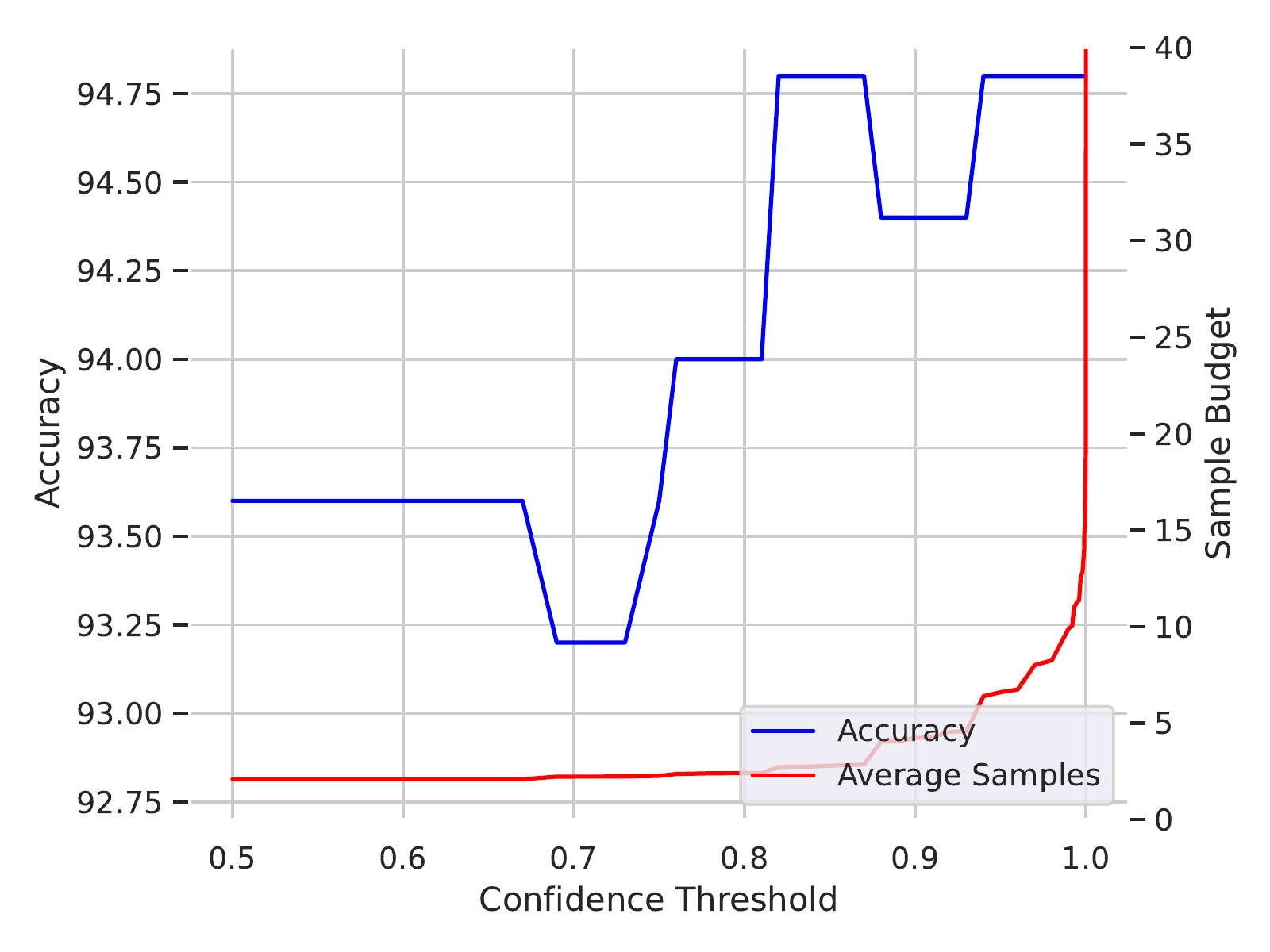}}
  \subfigure[\snarks]{\includegraphics[width=0.32\linewidth]{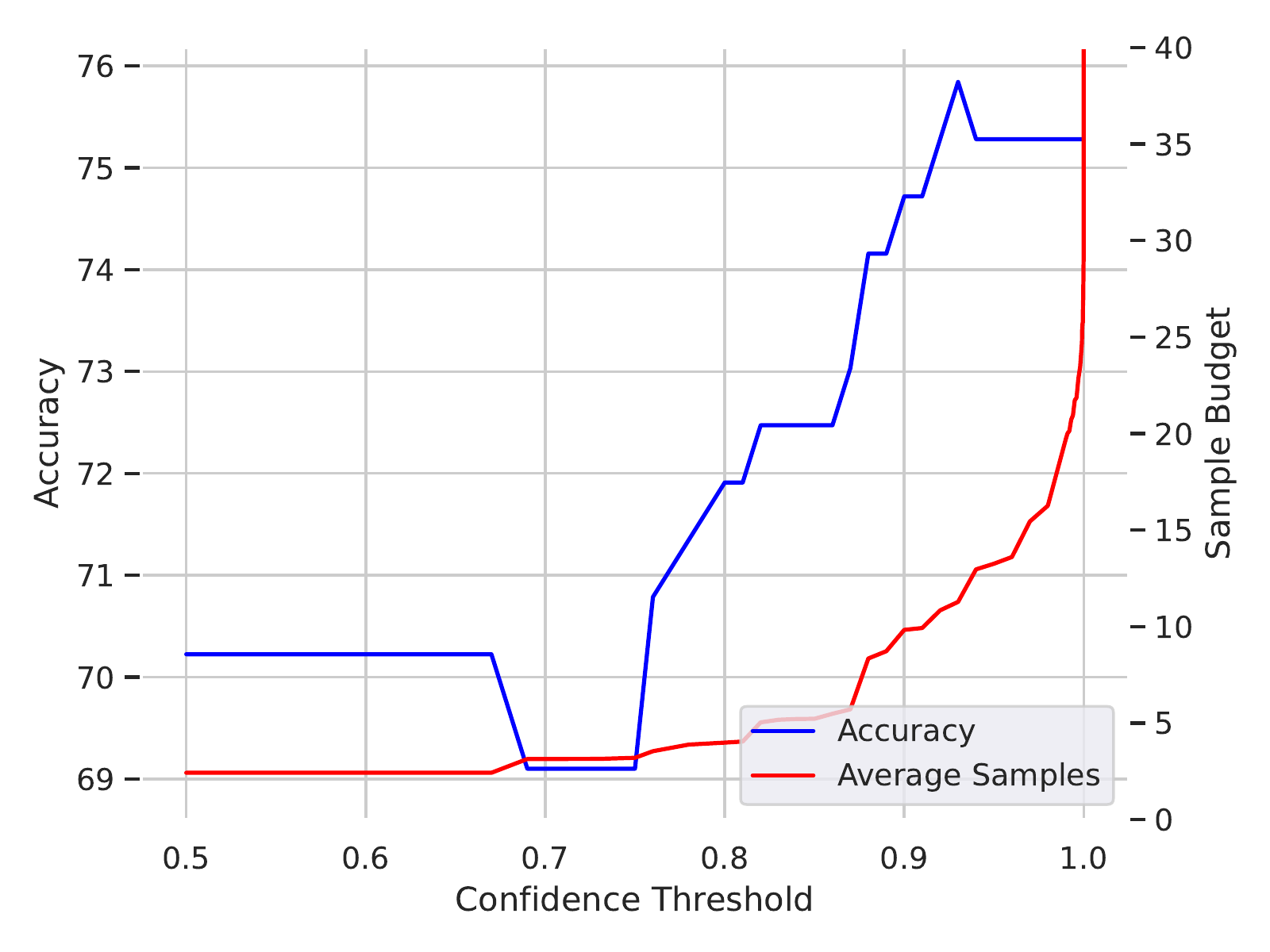}}
  \subfigure[\ruinnames]{\includegraphics[width=0.24\linewidth]{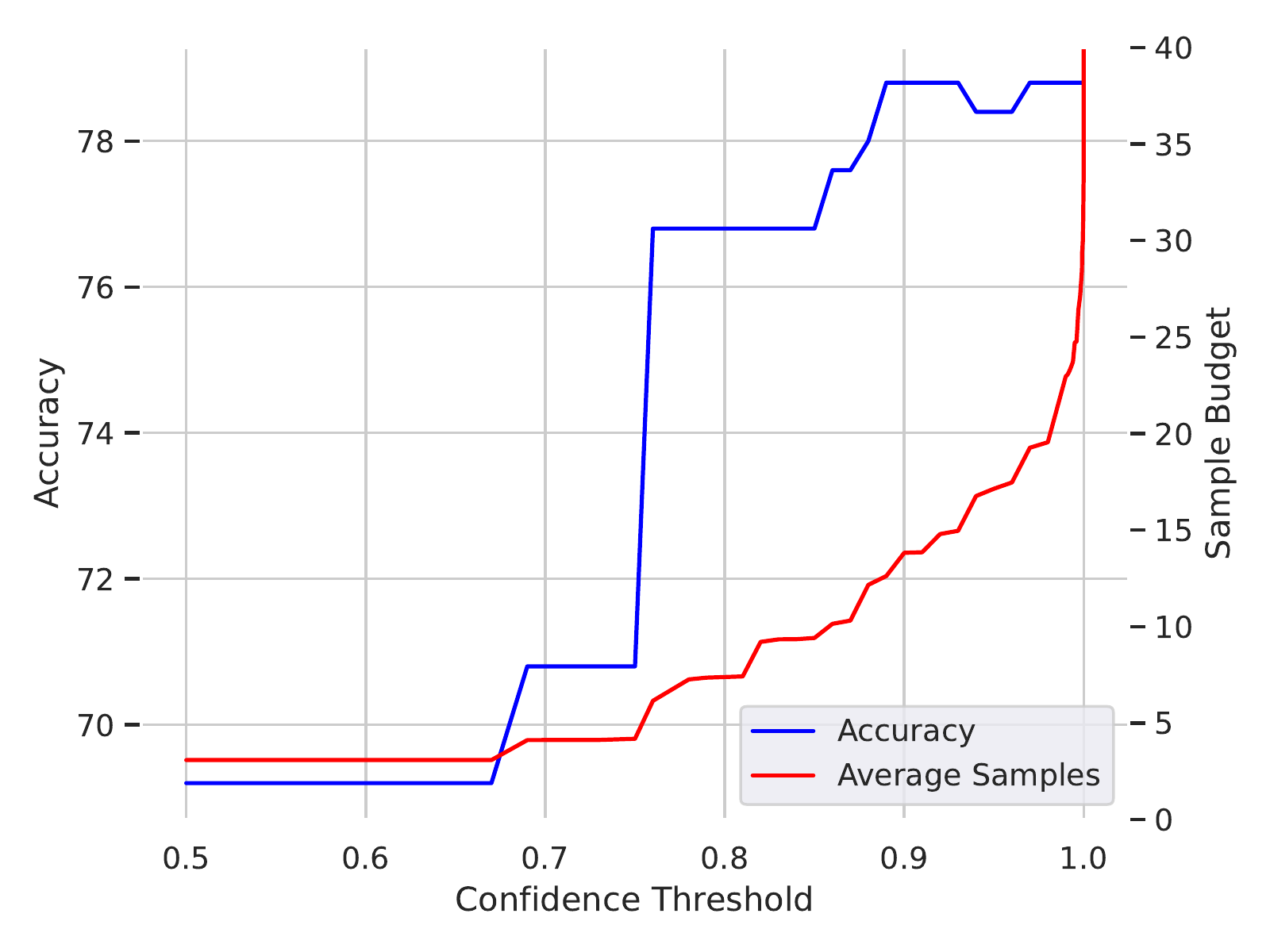}}
  \subfigure[\salienttranslation]{\includegraphics[width=0.24\linewidth]{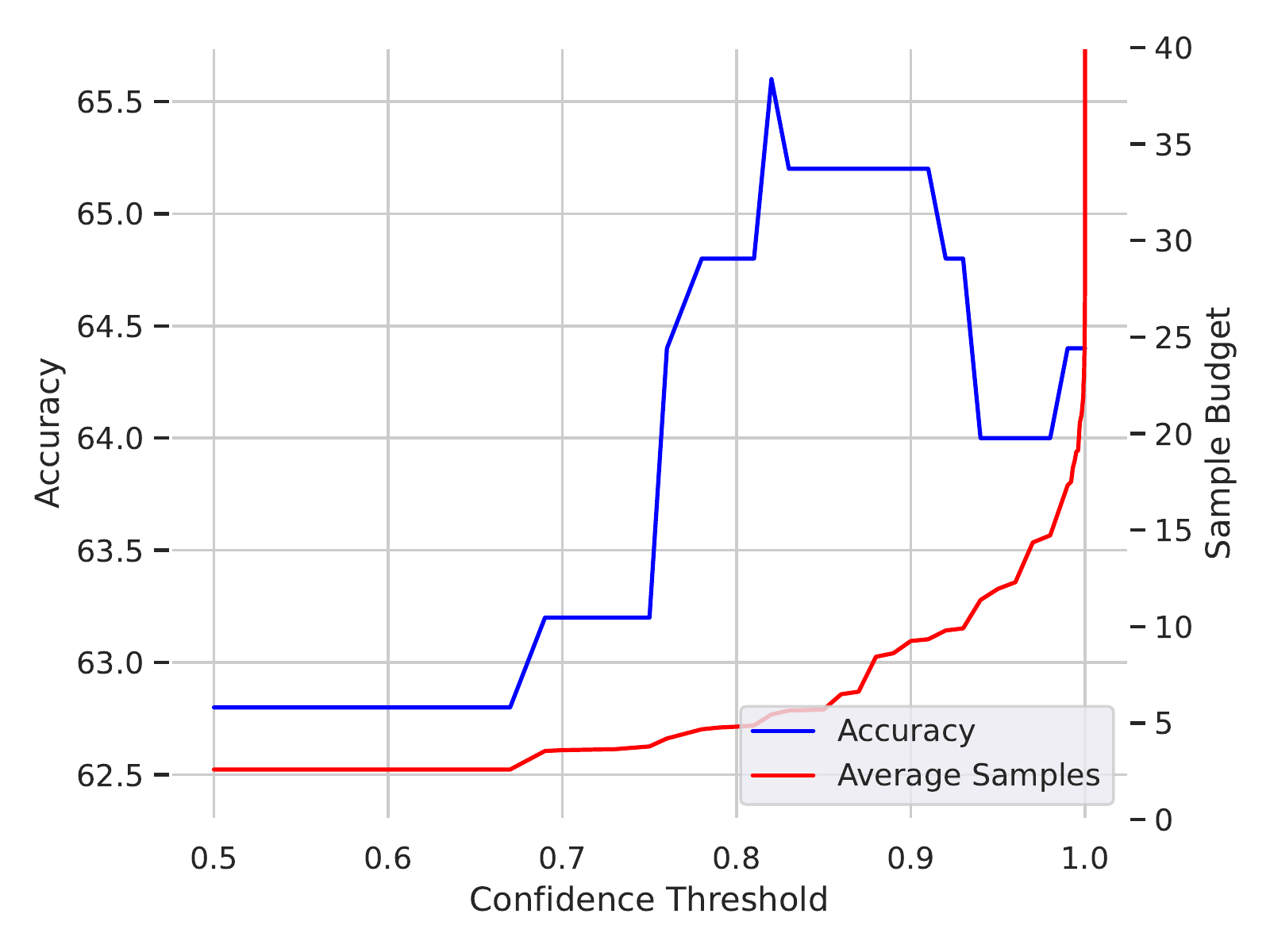}}
  \subfigure[\disambiguationqa]{\includegraphics[width=0.24\linewidth]{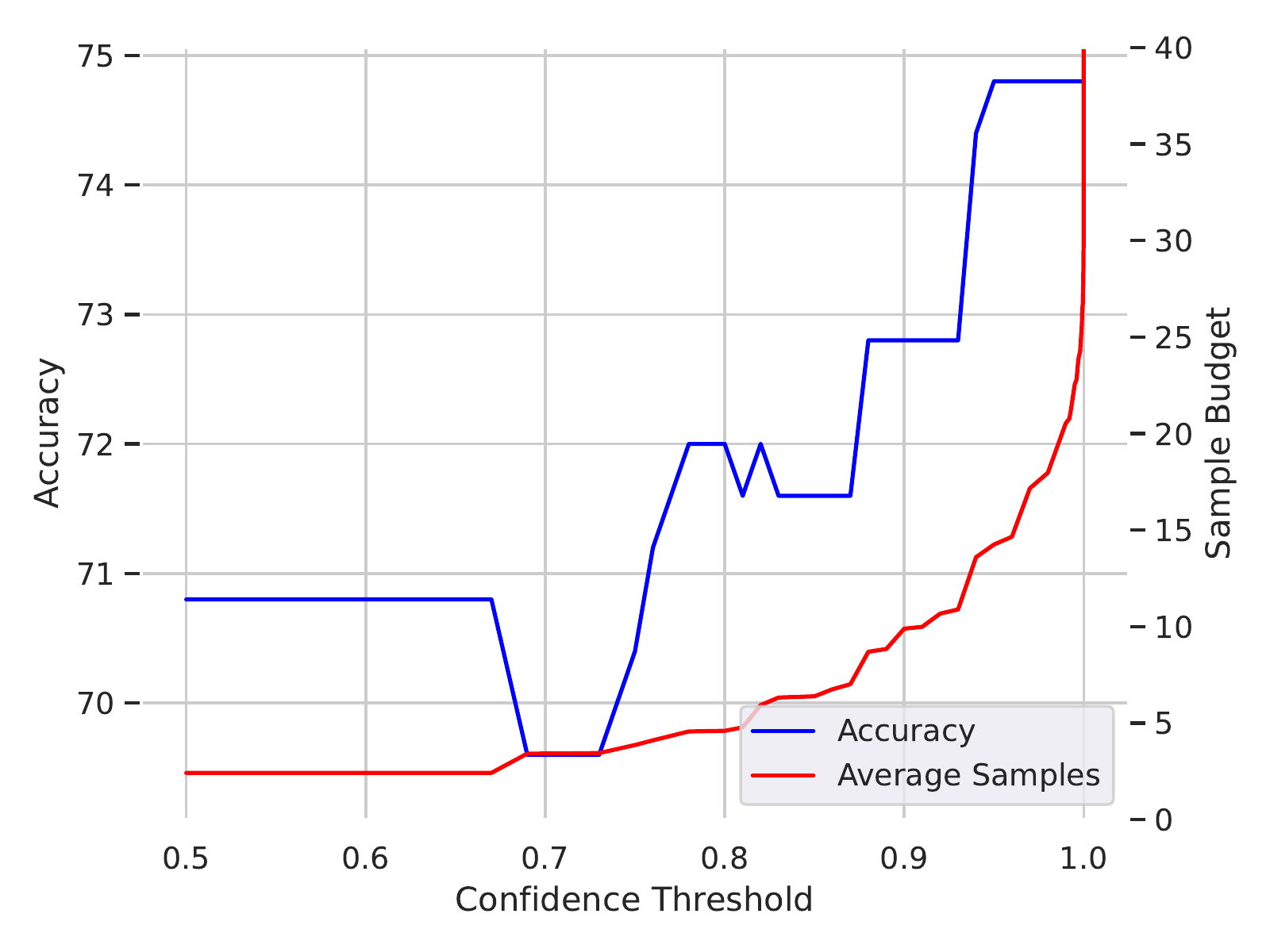}}
  \subfigure[\penguins]{\includegraphics[width=0.24\linewidth]{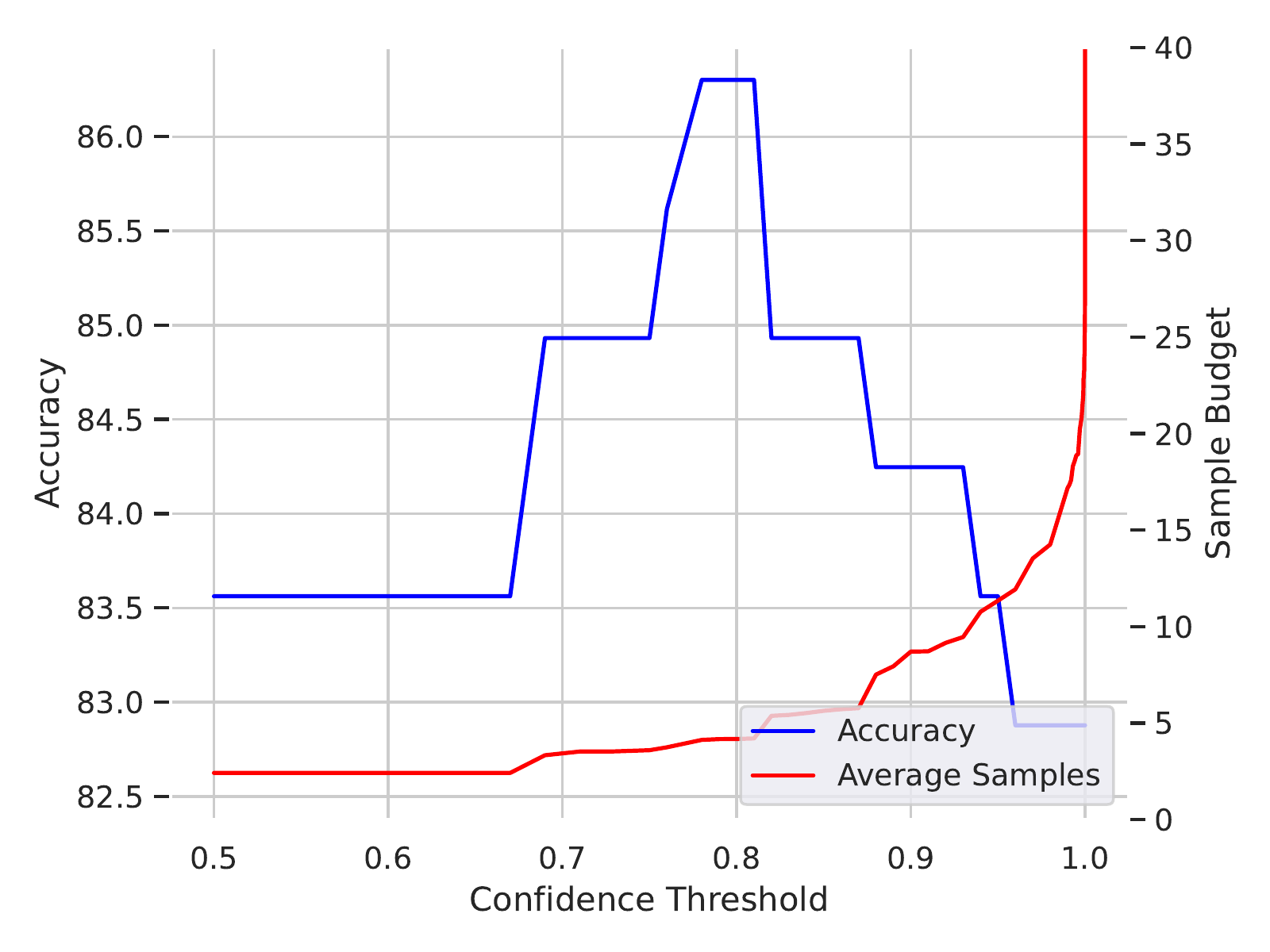}}

  \caption{Impact of Confidence Threshold ($C_{thresh}$) on \ours: As $C_{thresh}$ varies, the accuracy of \ours increases gradually, eventually plateauing. Initially, the average number of generations also increases gradually but then sharply climbs, reflecting the accuracy-confidence trade-off. The trend is observed almost consistently across all datasets.}
  \label{fig:appendix_cthreshanalysis}
\end{figure*}



\section{Derivation of \dirichlet \stopfunc}
\label{app:dirichletproof}


Consider for a given input ($I$), the model can generate one of $m$ distinct answers $A := \{a_1, a_2, \ldots a_m\}$. Define the probability of generating an answer given input as $p_i  := P(a_i \mid I) $. Now, consider an observation set ($O$) with counts of each of $a_i$ as $v_i$, such that $\sum_{i=1}^{m} v_i = n$. Now, without loss of generality, consider $p_1 > \max_{i=2}^{m} p_i$.
Now, based on Equation~\eqref{eq:probeq}, we need to find the probability:
$$P(p_1 > \max_{i=2}^{m} p_i \mid O)$$.

However, here the $p_i$s are latent variables, and only $O$ is available to us.
We next make the following \\
\textbf{Assumption 1:} The vector $\vec{p} = \{p_1, p_2 \ldots p_m\}$ is sampled from uniform distribution over $(m-1)$-simplex. \\
Thus, $p_1 = 1 - \sum_{i=1}^{m-1} p_i$.
Since the observation set follows a multinomial distribution with parameters $\vec{p}$, conditional joint probability distribution of $O$ given $\vec{p}$ can be written as:
$$P(O \mid \vec{p}) = \frac{n!}{\prod_{i=1}^{m} (v_i!)} \prod_{i=1}^{m} p_i^{v_i} = Dir(v_1 + 1, v_2 + 1 \ldots v_m + 1)$$, where $Dir$ represents the dirichlet distribution with $v_i + 1$, as its parameters.
Applying Baye's Rule,
$$P(\vec{p} \mid O) = \frac{P(O \mid \vec{p}) \cdot P(\vec{p})}{P(O)}$$.
Here $P(O)$ is a normalizing constant and can be omitted for computation. From Assumption 1, since $\vec{p}$ is sampled from uniform distribution, $$P(\vec{p}) = \prod_{i=2}^{m} dp_i$$
Thus conditional joint probability distribution of $\vec{p}$ given $O$ can be written as:
\begin{align}
P(\vec{p} \mid O) &= \text{Dir}(v_1 + 1, v_2 + 1, \ldots, v_m + 1) \nonumber \\
&\quad dp_m dp_{m-1} \ldots dp_2
\end{align}

Now we can integrate the above equation over a subset of $(m-1)$-simplex, such that $p_1 > \max_{i=2}^{m} p_i$. This gives us the equation:

\begin{equation}
\begin{aligned}
P&(p_1 > \max_{i=2}^{m} p_i \mid O) \\
&= \int_{0}^{1} \int_{\mathcal{S}(p_1')} P(\vec{p} \mid O) \\
&\qquad\qquad\qquad\qquad dp_2 \cdots dp_m dp_1', \\
\text{where}\\ \; \mathcal{S}(p_1') &= \{ (p_2, \ldots, p_m) \mid p_1' > \max_{i=2}^{m} p_i, \\
& \qquad\qquad\qquad\qquad\qquad \sum_{i=2}^{m} p_i = 1 - p_1' \}.
\end{aligned}
\label{eq:appendix_dirichprob}
\end{equation}

We note that the integration has no closed-form solution, and we use numerical approximation to compute the above integral.

\paragraph{Defining region of integration: $\mathcal{S}(p_1')$} 
Next, for computation of Equation~\eqref{eq:appendix_dirichprob}, we need to precisely calculate the limits of each integration such that they represent the region $\mathcal{S}(p_1')$. We do so by noting the following constraints on $p_i$: 1.) The $p_i = 0$ is valid $\forall 2 \leq i \leq m$, 2.) Given  $\{p_m, p_{m-1} \ldots p_{i+1}$ are fixed and in region  $\mathcal{S}(p_1')$, $p_i < \frac{1 - \sum_{j=i+1}^{m} p_j}{2}$ as else $p_i \geq p_1$ which is not allowed, 3.) Since $p_1 > \max_{j=i+1}^{m} p_j$, so $p_i < 1 - \sum_{j=i+1}^{m} p_j - \max_{j=i+1}^{m} p_j$, as else the $\vec{p}$, will lie outside the $(m-1)$-simplex, which is invalid. The first condition makes the lower limit for each integration 0, and the minimum of condition 2 and condition 3 gives the upper bound (limit) on each of the integrations.

\paragraph{\binomial{} \stopfunc}
Due to $m - 1$ dimensions integrations involved, with $m$ often getting larger than 10, computing Equation~\eqref{eq:appendix_dirichprob} is not efficient. Instead, we observe that establishing the majority of $p_1$ over the next largest probability, $p_2$, is sufficient for our purpose. Then, pdf simplifies to \binomial{} distribution with parameters $v_1 + 1, v_2 + 1$, and Equation~\eqref{eq:appendix_dirichprob} simplifies to:
\begin{equation}
    \int_{0}^{0.5} p_2^{v2} \cdot (1 - p_2)^{v_1} dp_{2}
\label{eq:appendix_betaeq}
\end{equation}

We use Scipy library in Python numerically compute the above equation.







\end{document}